\documentclass[lettersize,journal]{IEEEtran}
\usepackage{amsmath,amsfonts}
\usepackage{array}
\usepackage{textcomp}
\usepackage{stfloats}
\usepackage{url}
\usepackage{verbatim}
\usepackage{graphicx}
\usepackage{cite}
\usepackage{bm}
\usepackage{array}
\usepackage{amssymb}
\makeatletter
\let\NAT@parse\undefined
\makeatother
\usepackage{hyperref}  \hypersetup{hypertex=true,
            colorlinks=true,
            linkcolor=blue,
            anchorcolor=blue,
            citecolor=blue
            }
\usepackage{float}
\usepackage{subfigure}
\usepackage{color}
\usepackage{multirow}
\usepackage{amsthm}
\theoremstyle{remark}
\newtheorem{remark}{Remark}

\usepackage {amssymb}

\usepackage{algpseudocode}
\usepackage{footmisc}
\usepackage{threeparttable}
\usepackage{adjustbox}
\usepackage{booktabs}
\usepackage{utfsym}
\usepackage{makecell}
\definecolor{DarkRed}{rgb}{0.55, 0.0, 0.0}
\definecolor{DarkGreen}{rgb}{0.0, 0.5, 0.0}
\usepackage[ruled,vlined,linesnumbered]{algorithm2e}
\algnewcommand\algorithmicforeach{\textbf{for each}}
\algdef{S}[FOR]{ForEach}[1]{\algorithmicforeach\ #1\ \algorithmicdo}
\begin{document}
\title{\LARGE \bf
Stereo-Planner: Learning Stereo-to-Trajectory Planning for Zero-Shot Sim-to-Real Quadrotor Navigation
}
\author{
Hongming Shen$^{1 \dag}$, Junjie Lu$^{1,2\dag}$, Xun Chen$^{1}$, Wei Wang$^{1}$, Xinhang Xu$^{1}$,\\ Bailing Tian$^{2}$,~\IEEEmembership{Member,~IEEE}, Lihua Xie$^{1}$,~\IEEEmembership{Fellow,~IEEE}\thanks{This research is supported by the National Research Foundation (NRF),
Singapore, under the NRF Medium Sized Centre scheme (CARTIN).
}
\thanks{$^1$ The authors are with the School of Electrical and Electronic Engineering, Nanyang Technological University, Singapore 639798}
\thanks{$^2$ The authors are with the School of Electrical and Information Engineering, Tianjin University, Tianjin, China 300072.}
\thanks{$^\dag$ These authors contributed equally to this work.}}
\maketitle
\begin{abstract}
Autonomous flight in cluttered environments requires quadrotors to interpret visual observations and generate collision-free trajectories within limited onboard computing budgets.
However, existing navigation systems commonly rely on dedicated depth sensors or explicit depth reconstruction, while learning to generate trajectories directly from RGB images remains challenging due to the sim-to-real domain gap.
In this work, we propose Stereo-Planner, an end-to-end vision-to-trajectory framework that enables zero-shot sim-to-real planning from stereo RGB observations.
The key idea is to formulate trajectory generation as state- and scale-conditioned querying of latent stereo geometry.
A Scene Representor incorporates a pretrained vision grounding foundation model to extract geometric cues shared across simulated and real-world environments, supporting generalization across visual domains.
A lightweight Trajectory Decoder then translates these representations into trajectories through interleaved attention, separating trajectory-state interaction from scene-feature retrieval and incorporating the stereo baseline as a metric-scale cue.
With these designs, Stereo-Planner is trained entirely on simulated data by minimizing differentiable trajectory costs and deployed directly in real-world environments without fine-tuning.
To evaluate its generalization, we construct forest and industrial simulation environments with textures and lighting conditions that differ substantially from those encountered during training.
Benchmark results demonstrate strong cross-scene generalization, with a $95\%$ success rate in both environments at a commanded speed of $8\,\mathrm{m/s}$, and competitive performance against depth-based planners using only stereo RGB observations for scene perception.
Furthermore, through pruning and distillation of the Scene Representor, Stereo-Planner substantially reduces the mean onboard inference latency from $24.27\,\mathrm{ms}$ to $14.44\,\mathrm{ms}$ on an NVIDIA Orin NX while preserving the cross-domain consistency of its geometric representations, enabling autonomous flight at speeds approaching $8\,\mathrm{m/s}$ on a compact quadrotor.
The software, simulation environments, and hardware designs will be open-sourced at \url{https://github.com/Stereo-Planner/Stereo-Planner} to facilitate future research and applications.
\end{abstract}
\section{Introduction}
\begin{figure}
    \centering
    \includegraphics[width=\linewidth]{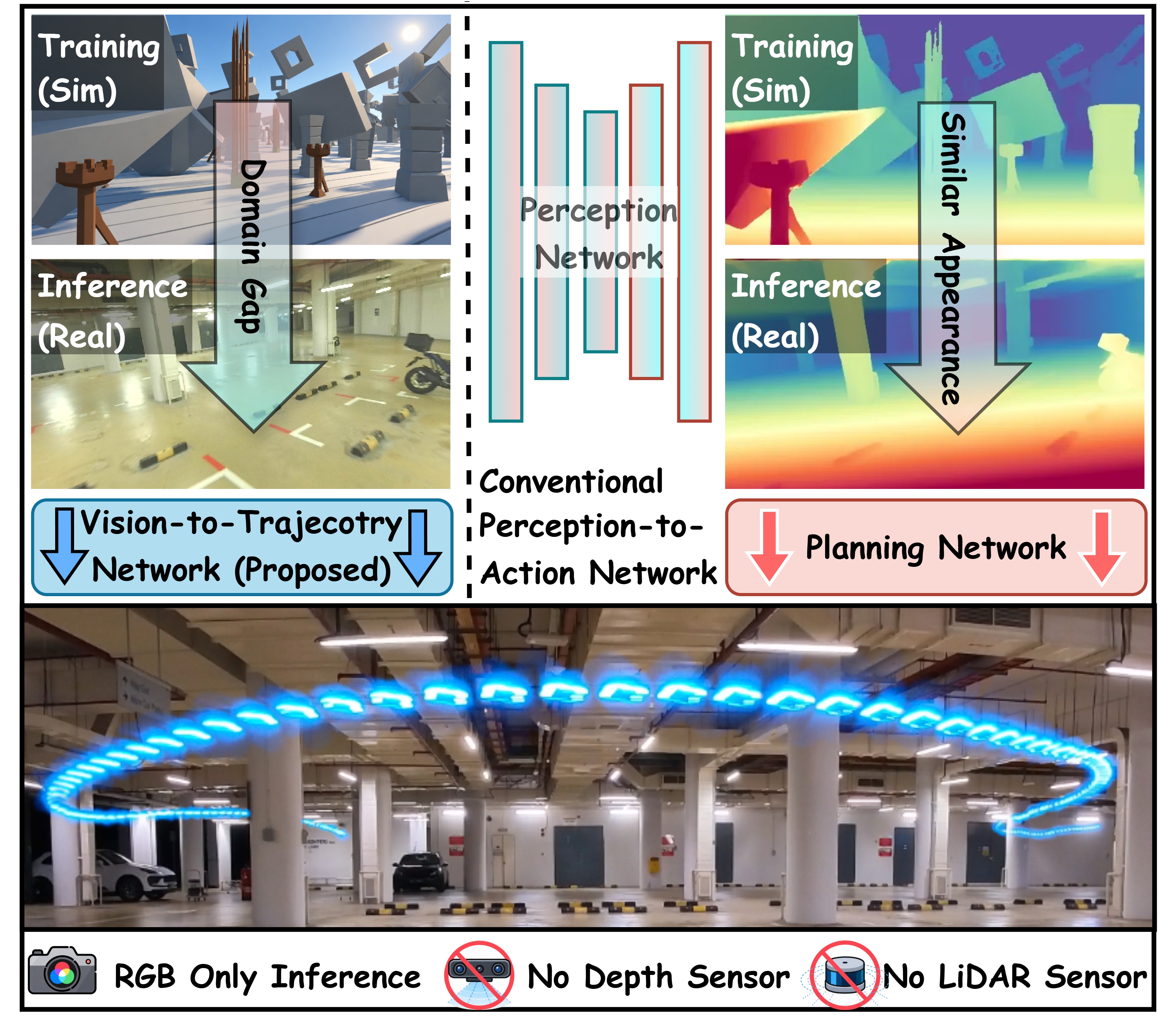}
    \vspace{-1.5em}
    \caption{Stereo-Planner: an end-to-end vision-to-trajectory framework that directly generates agile, collision-free trajectories from onboard stereo RGB observations and enables zero-shot sim-to-real transfer without explicit depth reconstruction or dedicated depth sensors.}
    \label{Fig: TileImage}
    \vspace{-1.5em}
\end{figure}
\IEEEPARstart{Q}{uadrotors}, among the most agile man-made aerial vehicles, embody humanity's aspiration to achieve bird-like agility in flight.
Their maneuverability enables them to navigate complex environments and access areas that are difficult for humans or ground robots to reach, supporting applications such as search and rescue\cite{10876078}, inspection\cite{7587351}, and aerial photography\cite{9998054}.
However, mastering agile flight in cluttered environments through manual control requires extensive pilot training and practice.
Recent advances in embodied artificial intelligence have enabled autonomous quadrotors to surpass expert human pilots in highly agile flight tasks, as corroborated in \cite{Kaufmann2023}. Nevertheless, such demonstrations often focus on acrobatic flight in partially known environments, where environmental priors, such as predefined reference trajectories and pre-acquired point cloud maps, are readily available.
Enabling quadrotors to learn to fly autonomously in previously unseen environments remains challenging, particularly in the following respects.

\textbf{End-to-End Vision-to-Trajectory Network:} Recent learning-based approaches have been explored to enable agile quadrotor flight in unknown and cluttered environments through imitation learning\cite{flightwild,Neural-Primitive}, privileged learning\cite{yopo,yopov2}, and differentiable Reinforcement Learning (RL)\cite{differentiablephysics}.
By learning implicit representations of the surrounding environment, these approaches directly generate trajectories from onboard sensory observations, establishing an end-to-end perception-to-action paradigm that bypasses the conventional sequential perception–mapping–planning pipeline\cite{doi:10.1126/scirobotics.abm5954,ren2025safety,LU2022107521}.
However, the aforementioned approaches still rely on depth information obtained through depth estimation algorithms\cite{doi:10.1126/science.968482,FoundationStereo,depthanything3} or dedicated depth sensors, retaining an intermediate geometric representation between visual observations and motion planning, as shown in Fig.~\ref{Fig: TileImage}.
The remarkable ability of birds to exploit binocular visual cues for agile flight and obstacle avoidance \cite{bhagavatula2011optic} raises a natural question: \textit{Can an autonomous quadrotor similarly translate onboard stereo observations directly into agile and collision-free trajectories?}
To this end, we propose Stereo-Planner, an end-to-end vision-to-trajectory framework that directly infers agile and collision-free trajectories from onboard stereo observations.

\textbf{Sim-to-Real Generalization:} As illustrated in Fig.~\ref{Fig: TileImage}, RGB images exhibit synthetic-to-real differences in texture and lighting that are less directly reflected in depth maps, making it challenging to transfer vision-to-trajectory policies trained in simulation to real-world environments.
Photorealistic simulation environments\cite{low2024sousvide,lv2026zeroshot}, constructed using novel view synthesis methods such as 3D Gaussian Splatting (3DGS)\cite{kerbl2023gaussian} and Neural Radiance Fields\cite{mildenhall2020nerf}, offer one route to narrowing the visual domain gap.
However, their reliance on scene-specific image acquisition and reconstruction makes it costly to build diverse training environments.
This data acquisition burden can constrain the scale and diversity of training datasets, potentially limiting the generalization of the resulting policies.
Meanwhile, advances in Vision Transformers (ViTs)\cite{ViT} and large-scale pretraining have enabled Visual Foundation Models (VFMs)\cite{DINO,oquab2023dinov2,depthanything3} to learn transferable representations from diverse datasets, including both synthetic and real-world imagery.
These representations offer the potential to mitigate synthetic-to-real appearance differences and improve feature robustness to illumination changes and motion blur.
Nevertheless, effectively transferring these capabilities to vision-based motion planning remains an open challenge.

\textbf{Real-Time Deployment on Compact Platforms:} A further challenge in integrating VFMs into motion planning is meeting the real-time requirements of agile flight, which calls for high-frequency planning updates (e.g., at 20~Hz) to respond promptly to surrounding obstacles.
However, the size, weight, and power constraints of compact quadrotors restrict onboard computing capacity, whereas the large parameter counts and high-dimensional representations of VFMs impose substantial computational demands.
To alleviate this computational burden, some approaches\cite{Cao-RSS-23,8594461,doi:10.1177/02783649251400382} offload motion planning to ground-based servers via tethered connections to meet real-time response requirements.
Such arrangements, however, are impractical in environments where external computing infrastructure is difficult to deploy.
Achieving low-latency trajectory inference within limited onboard computing budgets is therefore essential for autonomous operation in previously unknown environments.
The system-level challenge is to jointly design the network architecture, onboard deployment strategy, and hardware platform to enable a compact, fully autonomous, and open-source quadrotor.
\subsection{Related Work}
This section reviews the literature in relation to the aforementioned challenges.
\subsubsection{Vision-based Autonomous Navigation}
Owing to the compact size, lightweight nature, and cost-effectiveness of vision sensors, vision-based navigation has received increasing attention over the past decade.
Early methods \cite{depth_nav1,depth_nav2} performed obstacle avoidance directly in the depth space, but their performance was highly susceptible to noise and uncertainty in depth measurements.
To obtain more robust geometric representations, Euclidean Signed Distance Field (ESDF) maps\cite{8903279,8968199,oleynikova2017voxblox} have been widely adopted for quadrotor motion planning.
Specifically, \cite{7759784} and \cite{fast} represent quadrotor trajectories using piecewise polynomials and B-splines, respectively, and optimize them using ESDF gradients.
However, ESDF construction introduces non-negligible computational overhead.
CUDA-parallelized ESDF construction\cite{9782137,mppi_planner}, combined with gradient-free model predictive path integral control\cite{8558663}, has been employed to reduce the computational latency of mapping and motion planning.
Ego-Planner \cite{ego} further reduces planning latency by eliminating ESDF construction and directly deriving collision costs and gradients from obstacle information in the voxel map.

Despite the interpretability of this modular structure, which consists of perception, mapping, and planning, it introduces additional latency that limits high-speed agile flight. 
Neural networks offer an alternative by directly mapping visual observations to trajectories or control actions. RL methods \cite{Racing_SR,Whole_Body} have achieved impressive pixel-to-action control for drone racing, but their data-intensive training and slow convergence often limit them to task-specific scenarios. Imitation learning \cite{flightwild,deep_panther} provides a more efficient alternative by learning policies from expert demonstrations, achieving strong performance in obstacle avoidance. Differentiable simulation \cite{differentiablephysics} further alleviates the slow convergence of conventional RL and enables high-speed navigation. In contrast, \cite{yopo,yopov2} directly leverage environmental gradients for training, eliminating the need for expert annotations and online rollouts and substantially simplifying the training process. Beyond end-to-end trajectory prediction, learning-based methods have also been explored to address planning aspects that are difficult to model analytically, such as adaptive speed selection \cite{Mavrl} and learned time allocation \cite{time_allo}.

\subsubsection{Sim-to-Real Transfer}
Compared with conventional modular navigation pipelines with explicit and interpretable structures, learning-based approaches often face challenges in sim-to-real transfer.
Domain randomization has been widely adopted to improve transferability by introducing variations in the simulated environment. 
In particular, control-based methods \cite{geles2024demonstrating,dynamics_domain_random} generally incorporate dynamics variations and randomized external disturbances during training, thereby reducing reliance on an idealized nominal dynamics model. 
For vision-based navigation \cite{loquercio2020deep}, appearance randomization, including variations in scene textures, colors, and illumination, has been employed to improve the robustness of learned policies in real-world environments. 
Additionally, some approaches \cite{yopo,yopov2} train policies across a diverse set of randomized environments to learn more generalizable navigation behaviors for unseen real-world scenarios. 
Furthermore, domain adaptation provides another approach to improving generalization. It aims to reduce feature discrepancies across domains during training, enabling the network to learn more robust representations.
In \cite{xing2021domain}, images with the same content but different illumination and weather conditions are jointly fed into the network to encourage similar feature representations and consistent predictions. 
In \cite{DomainAdaptation,DomainAdaptation2}, an adversarial learning strategy is adopted to make the learned features indistinguishable between the simulated and real domains.
Some approaches aim to improve the realism of training data and reduce the gap between the training and deployment distributions. 
For example, \cite{flightwild} uses the Semi-Global Matching algorithm to generate depth images, simulating occlusions and measurement imperfections of real depth cameras. 
The YOPO series\cite{yopo,yopov2} and DiffPhysDrone\cite{differentiablephysics} use low-resolution images to reduce discrepancies in fine-grained depth structures between simulation and the real world. 
In contrast, \cite{flow1,flow2} use optical flow instead of RGB images as input, which better captures motion cues while avoiding the appearance gap between simulated and real RGB images. 
In comparison, \cite{low2024sousvide,lv2026zeroshot} use 3DGS to render realistic RGB images, providing photorealistic simulation and enriching scene diversity through variations in weather and illumination.

\subsubsection{Compact Autonomous Quadrotors}

Compact autonomous quadrotors must integrate sensing, onboard computing, and flight control under stringent size, weight, and power constraints. 
Early integrated systems, such as the platform developed by \cite{mohta2018fast}, combined stereo visual state estimation with a mechanically actuated nodding 2D LiDAR for three-dimensional mapping in Global Navigation Satellite System (GNSS)-denied environments.
Benefiting from the accurate, long-range, and direct depth measurements provided by LiDAR, recent studies have demonstrated highly robust state estimation \cite{FASTLIO2,cte-mlo,UniLGL} and high-performance motion planning \cite{9981518,10970076,8593826} for autonomous quadrotor flight.
With the recent trend toward lighter and more compact LiDAR sensors, the size of LiDAR-equipped autonomous quadrotors has been progressively reduced, from the DJI M600 platform equipped with a Velodyne VLP sensor in \cite{8593826}, to a $350\,\mathrm{mm}$-wheelbase quadrotor carrying an Ouster LiDAR in \cite{minco}, and further to a $280\,\mathrm{mm}$ platform equipped with a Livox Mid-360 in \cite{ren2025safety}.
Nevertheless, mechanically scanned LiDARs generally rely on moving components to steer or emit laser beams, making them more susceptible to impact and mechanical damage than passive cameras. 
This limitation is particularly relevant for quadrotors, which are inherently prone to crashes and collisions, leading to higher maintenance and replacement costs for LiDAR-based systems. Consequently, mainstream commercial aerial platforms, such as DJI Mavic, Skydio, and Autel drones, more commonly adopt stereo-vision-based perception solutions.
Agilicious\cite{foehn2023agilicious}, for example, combines an Intel RealSense T265 for real-time state estimation with an Intel RealSense D400 series RGB-D camera for dense depth perception, enabling learned agile flight in complex environments.
Inspired by this system design, YOPO\cite{yopo} adopts VINS-Fusion\cite{vins-fusion} for state estimation and introduces an end-to-end planning network, achieving a more balanced allocation of onboard computational resources while avoiding the redundant T265-plus-D400 sensing configuration used in Agilicious\cite{foehn2023agilicious}.
DiffPhysDrone\cite{differentiablephysics} further reduces onboard computational requirements by removing the dependence on explicit visual odometry and enabling real-time planning inference even on a lightweight Mango Pi embedded computer.
Nevertheless, these systems still rely on depth cameras, whose size and weight remain non-negligible for compact aerial platforms.
This motivates compact RGB-only aerial platforms such as OmniNxt \cite{liu2024omninxt}, which employs only RGB cameras and adapts EGO-Planner for autonomous navigation in cluttered environments.
However, EGO-Planner\cite{ego} still requires an online occupancy representation of the surrounding environment.
To provide such geometric information, OmniNxt integrates Omni-VINS and Omni-Depth for localization and dense mapping through coordinated hardware and software design. These perception modules, particularly online dense depth recovery, impose substantial onboard computational demands.

\begin{table*}[!t]
\centering
\caption{Comparison of autonomous quadrotor navigation methods and platforms.}
\label{tab: related_work}
\begingroup
\newcommand{\capyes}{{\color{DarkGreen}\usym{1F5F8}}}
\newcommand{\capno}{{\color{DarkRed}\usym{2717}}}
\setlength{\tabcolsep}{2pt}
\renewcommand{\arraystretch}{1.2}
\scriptsize
\begin{threeparttable}
\begin{tabular}{l||*{11}{c}}
\hline\hline
Capability &\makecell{YOPO V2\\\cite{yopov2}} &\makecell{DiffPhysDrone\\\cite{differentiablephysics}} &\makecell{Agile\\\cite{flightwild}} &\makecell{Depth Transfer\\\cite{DomainAdaptation2}} &\makecell{Pixel Motion\\\cite{flow1}} &\makecell{Relightable\\\cite{lv2026zeroshot}} &\makecell{Neural-Primitive\\\cite{Neural-Primitive}} &\makecell{SUPER\\\cite{ren2025safety}} &\makecell{Ego-Planner V2\\\cite{doi:10.1126/scirobotics.abm5954}} &\makecell{OmniNxt\\\cite{liu2024omninxt}} &\makecell{Stereo-Planner\\(Ours)}\\\hline
End-to-End &\capyes &\capyes &\capyes &\capyes &\capyes &\capyes &\capyes &\capno &\capno &\capno &\capyes\\
Sensor Input &RGB-D &RGB-D &RGB-D &RGB-D &Mono-RGB &Mono-RGB &LiDAR &LiDAR &LiDAR/RGB-D &Stereo-RGB &Stereo-RGB\\
Sim-to-Real &Depth &Depth &Depth &Depth &Optical Flow &3DGS &PointCloud &- &- &- &VFM\\
Vision-to-Trajectory\tnote{1} &\capno &\capno &\capno &\capno &\capno &\capyes &\capno &\capno &\capno &\capno &\capyes\\
Open-source Software &\capyes &\capyes &\capyes &\capyes &\capno &\capno &\capno &\capyes &\capyes &\capyes &\capyes\\
Open-source Hardware &\capyes &\capno &\capyes &\capno &\capno &\capno &\capno &\capyes &\capno &\capyes &\capyes\\
\hline\hline
\end{tabular}
\begin{tablenotes}[flushleft]
\scriptsize
\item[1] Vision-to-Trajectory denotes direct RGB-image-to-trajectory prediction without explicit depth recovery or mapping.
\end{tablenotes}
\end{threeparttable}
\endgroup
\vspace{-1.5em}
\end{table*}

\subsection{Motivation and Contributions}
\begin{figure}[!t]\centering
\includegraphics[width=\linewidth]{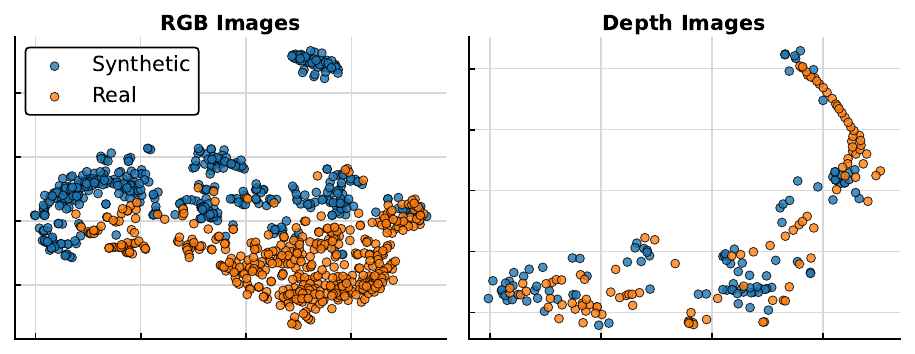}
\vspace{-1.5em}
\caption{t-SNE\cite{tsne} visualization of the synthetic and real-world RGB images shown in Fig.~\ref{Fig: TileImage} and their corresponding depth images, illustrating the sim-to-real domain gap in each modality.
}\label{Fig: tsne-Motivation}
\vspace{-1em}
\end{figure}

In view of the aforementioned analysis, as summarized in Table~\ref{tab: related_work}, existing quadrotor navigation systems commonly rely on conventional perception and mapping pipelines\cite{ren2025safety,liu2024omninxt,ego}, dedicated depth sensors\cite{yopov2,differentiablephysics,flightwild,Neural-Primitive}, or intermediate perception algorithms such as depth estimation\cite{liu2024omninxt} and optical flow\cite{flow1}. Direct end-to-end trajectory generation from RGB observations remains less explored. One underlying challenge is the sim-to-real domain gap: RGB images are sensitive to scene appearance and illumination, whereas depth observations provide a more consistent geometric representation across domains, as illustrated in Fig.~\ref{Fig: tsne-Motivation}.
Recent VFMs have exhibited strong cross-domain generalization in visual representation, but their use in motion planning remains largely unexplored.
\textit{This motivates us to leverage a VFM to directly reduce the sim-to-real appearance gap at the representation level, enabling zero-shot and RGB-only trajectory planning in real-world environments without requiring real-to-sim reconstruction or real-world fine-tuning.}
From a deployment perspective, although numerous autonomous flight frameworks have been open-sourced, with several also providing hardware reference designs, most existing systems achieve onboard autonomy through either conventional perception-mapping-planning pipelines\cite{ren2025safety,liu2024omninxt,ego} or lightweight neural networks\cite{yopov2,differentiablephysics,flow1,lv2026zeroshot,Neural-Primitive}, such as convolutional neural networks.
The practical deployment of VFMs on resource-constrained aerial edge-computing platforms, however, remains largely unexplored. Motivated by this gap, we present a complete autonomous flight solution spanning algorithm design, software implementation, onboard deployment, and hardware system integration.
The main contributions of this paper are listed as follows:
\begin{itemize}
    \item \textbf{Stereo-RGB Inference}: We propose an end-to-end vision-to-trajectory network that directly maps stereo RGB observations to feasible trajectories. Specifically, Stereo-Planner uses a geometry-aware encoder to extract implicit scene geometry from stereo RGB images and an interleaved-attention decoder to generate regularized trajectory candidates. This design enables RGB-only trajectory inference without explicit depth recovery or redundant depth/LiDAR sensors.
    \item \textbf{Zero-Shot Sim-to-Real Transfer}: Following the paradigm shift brought by VFMs in visual perception, Stereo-Planner explores the feasibility of leveraging geometric representations directly from the pretrained encoder of a vision grounding foundation model for trajectory planning, bypassing its heavy depth prediction head.
    Furthermore, we investigate structured pruning and knowledge distillation to compress the VFM-based Scene Representor, substantially reducing onboard inference latency while preserving the cross-domain consistency of its geometric representations. To the best of our knowledge, Stereo-Planner is the first quadrotor motion planning framework to leverage a vision grounding foundation model for zero-shot sim-to-real agile flight, enabling direct real-world deployment through simulation-only training.
    \item \textbf{Complete Autonomous Flight System}: Stereo-Planner is a full-fledged Micro Aerial Vehicle (MAV) locomotion framework that open-sources the complete ROS2\cite{ros2} algorithmic pipeline, simulation environments, and hardware design. Evaluations across both simulated and real-world scenarios demonstrate its generalization capability and practical applicability, while benchmark comparisons show that Stereo-Planner achieves planning performance competitive with methods that require explicit depth input, while using only stereo RGB observations. Moreover, zero-shot high-speed flight approaching $8\,\mathrm{m/s}$ in real-world environments further demonstrates its strong performance and deployment practicality.
\end{itemize}

\section{Problem Formulation}
Consider a trajectory $\mathbf{p}(t)\in \mathbb{R}^3$, $t \in \left[0,T\right]$, represented by a single-segment time-parameterized polynomial:
\begin{equation}
{\bf{p}}(t) = {\bf{C}^\top {\bm\beta}}(t)
\label{eq_1}
\end{equation}
where ${\bf C}\in\mathbb{R}^{2o\times 3}$ is the polynomial coefficient matrix, and ${\bm\beta}=\left[1,t,\ldots,t^{2o-1}\right]^\top$ is the polynomial basis vector.
The coefficient matrix ${\bf C}$ can be solved in closed form from the current and terminal states of the trajectory:
\begin{equation}
{\bf{C}} = {{\bf{M}}^{ - 1}}\left[ {\begin{array}{l}
{{\bf{s}}(0)}\\
{{\bf{s}}(T)}
\end{array}} \right]
\label{eq_2}
\end{equation}
where $\mathbf{M} \in \mathbb{R}^{2o \times 2o}$ is a constant coefficient mapping matrix, and ${\bf{s}}(0)$ and ${\bf{s}}(T)$ denote the current and terminal states, respectively, each defined up to the $(o-1)$-th derivative.

In this work, we set $o=3$, so that the resulting fifth-order polynomial trajectory ${\bf p}(t)$ corresponds to a minimum-jerk trajectory.
A neural network $\Phi$ is designed as a mapping function that infers the terminal state $\mathbf{s}(T)$ from the stereo camera observations $\left\{I_0, I_1\right\}$ and the current state ${\bf{s}}(0)$:
\begin{equation}
    \Phi: \left\{I_0, I_1, \mathbf{s}(0)\right\} \rightarrow \mathbf{s}(T)
\end{equation}
Substituting the terminal state predicted by $\Phi$ into (\ref{eq_2}) fully parameterizes the polynomial trajectory defined in (\ref{eq_1}).
Therefore, the key to motion planning in this work is to design a terminal-state prediction network that enables the quadrotor to safely navigate toward the target.

\section{The Stereo-Planner Model}
\begin{figure*}
    \centering
    \includegraphics[width=\linewidth]{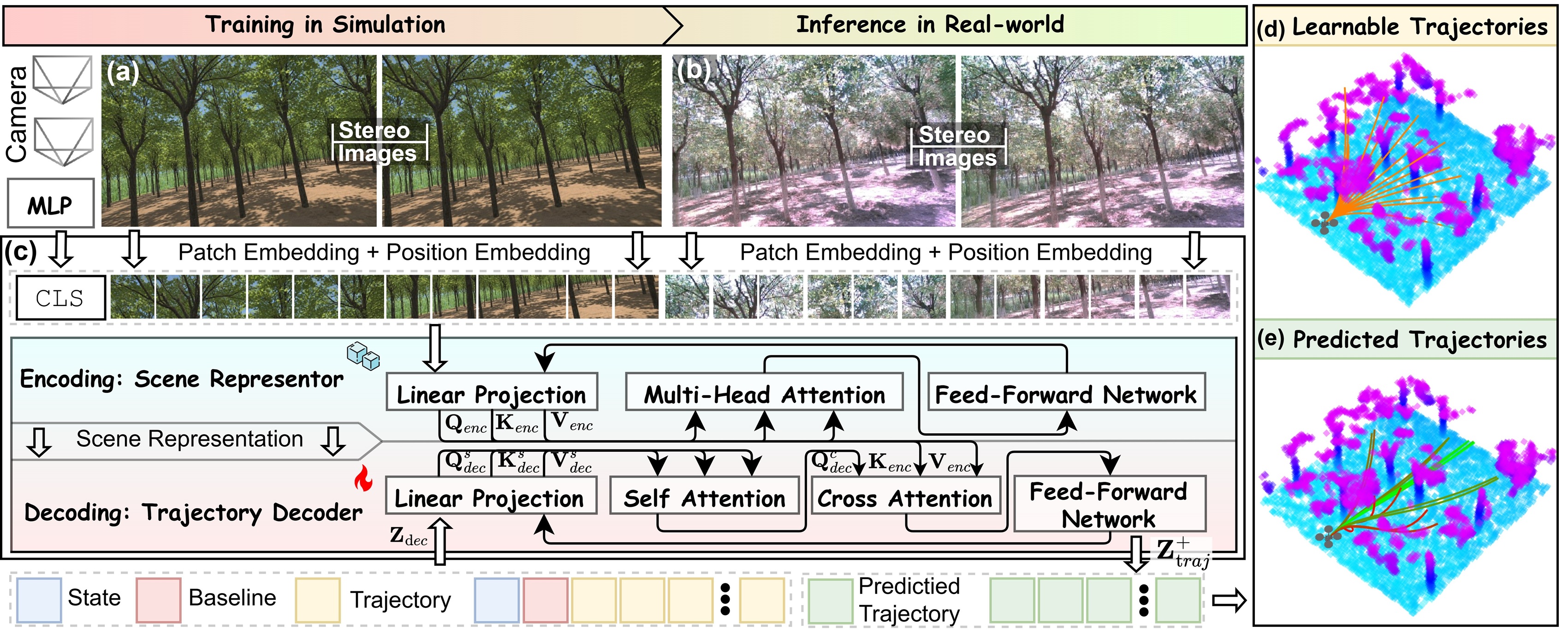}
    \vspace{-1em}
    \caption{Network architecture of Stereo-Planner. The encoder-decoder design in (c) bridges the visual-domain gap between the simulated and real images shown in (a) and (b). Conditioned on the learned scene representation, the decoder refines the learnable trajectory (d) to infer the collision-free trajectory (e).}
    \label{Fig: NetworkArchitecture}
    \vspace{-1em}
\end{figure*}

The proposed Stereo-Planner, illustrated in Fig.~\ref{Fig: NetworkArchitecture}, follows a simple encoder--decoder architecture inspired by recent vision grounding models\cite{D4RT,depthanything3,wang2025vggt}.
The encoder, referred to as the \emph{Scene Representor}, constructs a holistic representation of the surrounding environment.
Specifically, it extracts dense stereo correspondences from the left and right images by leveraging cues from the camera extrinsics, producing a geometry-aware scene representation.
A lightweight decoder, referred to as the \emph{Trajectory Decoder}, then queries the scene representation using the parametric trajectory representation together with the current MAV state and predicts the trajectory for autonomous flight.

\subsection{Scene Representor}\label{Sec: Scene Representor}
Stereo-Planner adopts the State-of-the-Art (SOTA) vision grounding model Depth Anything V3 (DA3) \cite{depthanything3} as the foundation of our Scene Representor.
Explicitly recovering high-quality metric depth, however, typically requires a large metric-depth model such as DA3-Large, introducing considerable computational overhead and system complexity.
We therefore discard the depth prediction head and retain only the pretrained DA3 encoder to extract geometry-aware scene representations, which are directly consumed by the Trajectory Decoder defined in Section~{\ref{Sec: Trajectory Decoder}} for motion planning.
\begin{equation}
{{\bf{Z}}_e} = {{\bf{\tau}}_e}\left( {{\mathbb{I}},{{\bf{T}}_\text{ext}},{\bf{K}}} \right)\label{Eq: Scene Representor}
\end{equation}
where $\mathbf{Z}_e \in \mathbb{R}^{(\rho+1) \times D}$ denotes a sequence comprising $\rho=2\left\lfloor {\frac{H}{P}} \right\rfloor \times \left\lfloor {\frac{W}{P}} \right\rfloor$ image tokens and one classification token, each with embedding dimension $D$; $\left\lfloor \cdot \right\rfloor$ denotes the floor operation; the encoder $\boldsymbol{\tau}_e(\cdot)$ is a Transformer with within-view and cross-view attention, initialized from the pretrained DA3 encoder to establish dense correspondences between the stereo images; $\mathbb{I}=\{\mathbf{I}_0,\mathbf{I}_1\}$ denotes the stereo image pair, while $\mathbf{T}_{\mathrm{ext}}\in \mathrm{SE}(3)$ and $\mathbf{K}\in \mathbb{R}^{3\times3}$ denote the extrinsic and intrinsic matrices of the stereo camera, respectively.

\begin{figure}
\centering	
\subfigure[Token-wise features from the Scene Representor and DINO V2 for the simulation (Fig.~\ref{Fig: NetworkArchitecture}{\color{blue}(a)}) and real-world (Fig.~\ref{Fig: NetworkArchitecture}{\color{blue}(b)}) images, with cosine similarity maps between their corresponding scene tokens.]{
    \includegraphics[width=0.95\linewidth]{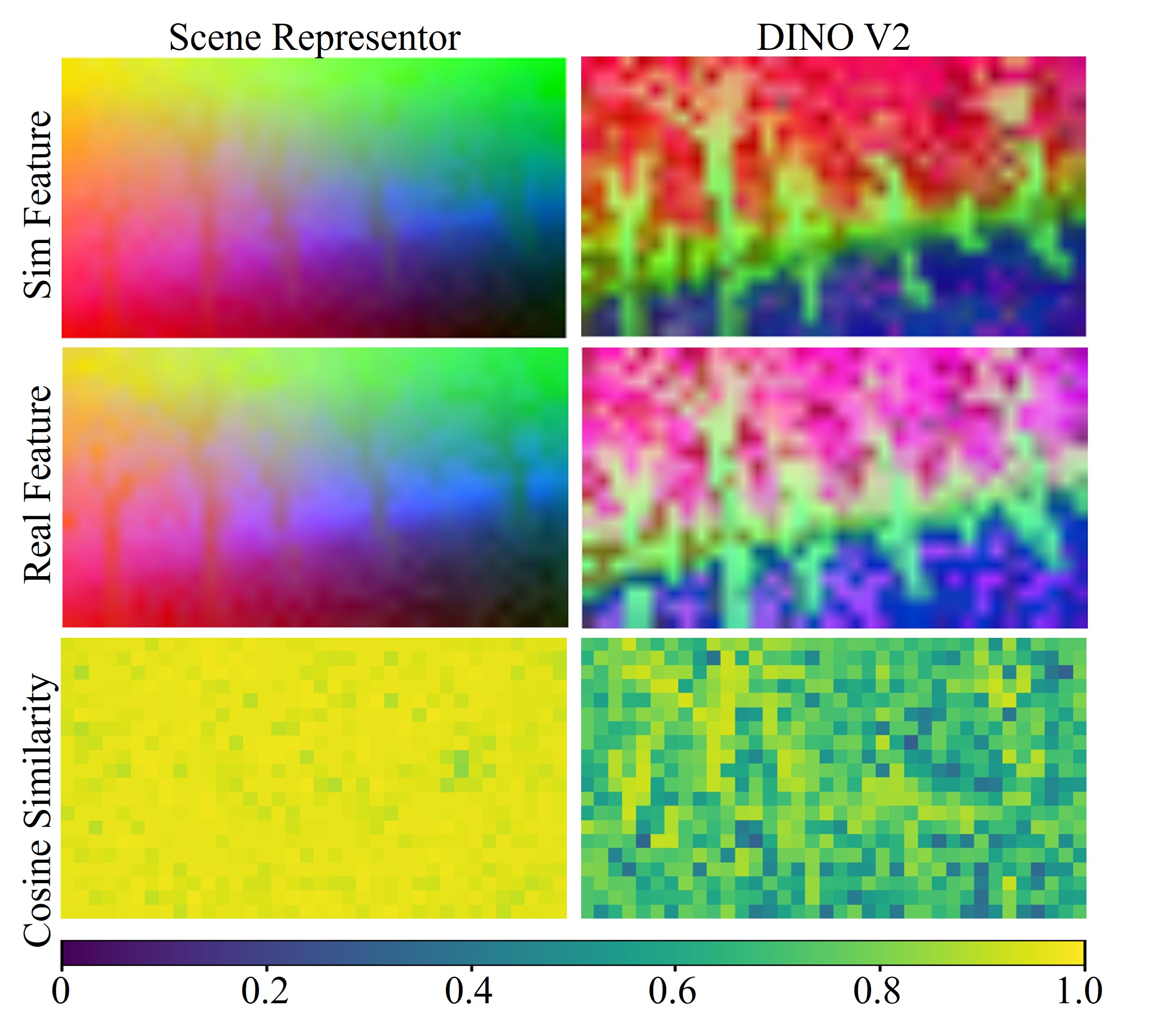} \label{Fig: Sim2RealFeatMap}
}
\subfigure[Token-wise cosine similarity distribution]{
    \includegraphics[width=0.95\linewidth]{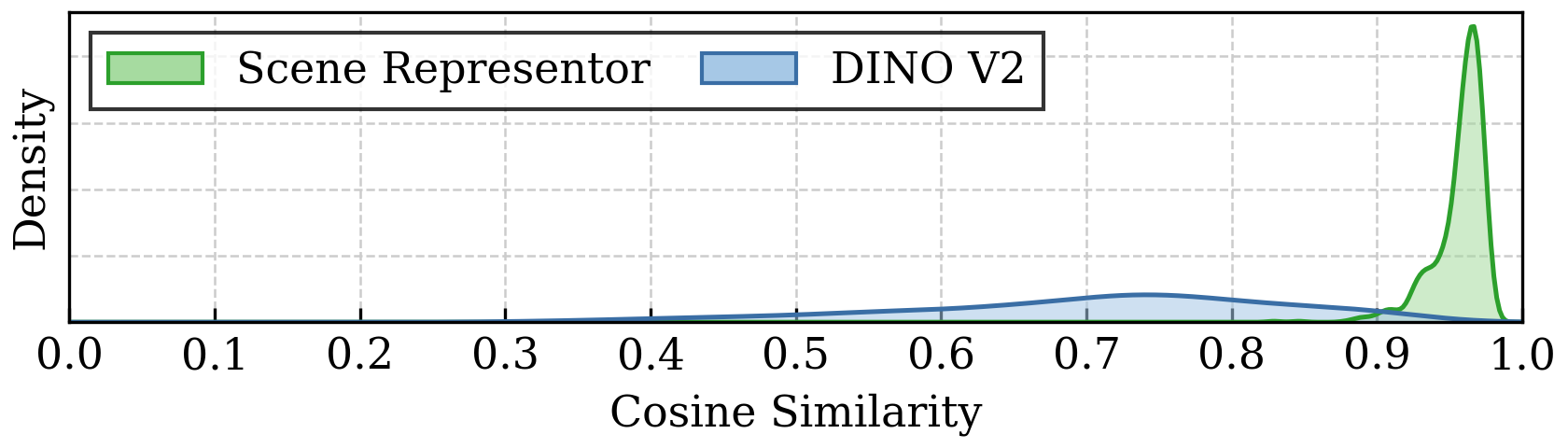} \label{Fig: Sim2RealSimilarityDis}
}
\caption{Demonstration of sim-to-real transfer.}
\label{Fig: Sim2Real}
\vspace{-1em}
\end{figure}

In this work, the Scene Representor defined in (\ref{Eq: Scene Representor}) is designed to provide a unified scene representation across simulated and real-world domains.
To investigate whether the vision grounding foundation model can help bridge the sim-to-real gap, we present a toy example in Fig.~\ref{Fig: Sim2Real} that compares the feature representations extracted from the simulated and real-world images in Fig.~\ref{Fig: NetworkArchitecture}{\color{blue}(a)} and Fig.~\ref{Fig: NetworkArchitecture}{\color{blue}(b)} using our Scene Representor and DINO V2 \cite{oquab2023dinov2}, a SOTA ViT-based VFM.
As illustrated by the token-wise feature maps in Fig.~\ref{Fig: Sim2RealFeatMap}, the proposed Scene Representor, initialized from the encoder of DA3, places greater emphasis on geometry-aware features that are more relevant to downstream planning, whereas the general-purpose ViT encoder DINO V2 focuses primarily on texture and semantic appearance cues.
Consequently, for the real-world image and its corresponding simulated counterpart shown in Fig.~\ref{Fig: NetworkArchitecture}{\color{blue}(b)} and Fig.~\ref{Fig: NetworkArchitecture}{\color{blue}(a)}, the geometric features extracted by the proposed Scene Representor exhibit significantly stronger sim-to-real consistency than those extracted by the SOTA ViT encoder DINO V2, as shown in the cosine similarity map in Fig.~\ref{Fig: Sim2RealFeatMap}.
Specifically, as shown in Fig.~\ref{Fig: Sim2RealSimilarityDis}, the token-wise cosine similarities between simulated and real-world scene tokens of the proposed Scene Representor are predominantly distributed above $0.9$.
Benefiting from the above properties, the proposed Stereo-Planner can be trained entirely on simulated data and directly deployed for inference in real-world environments.

\begin{remark} \label{rem: encoder distillation}
    \textbf{Scene Representor Distillation:} 
    \textit{In this work, the Scene Representor adopts the 12-layer ViT-S encoder of DA3\cite{depthanything3} as its backbone to provide a zero-shot unified scene representation for the MAV.
    Considering that low-level trajectory planning requires substantially higher inference frequencies than high-level tasks such as vision-language navigation\cite{vla_an}, manipulation\cite{10339889}, and exploration\cite{10816079}, we further investigate a distillation strategy for the Scene Representor, making it more suitable for real-time onboard trajectory planning. The details are presented in Appendix~\ref{app: Distill}.}
\end{remark}

\subsection{Trajectory Decoder} \label{Sec: Trajectory Decoder}
The Trajectory Decoder efficiently maps the scene representation to multiple candidate trajectories using a lightweight set of queries constructed from the parametric trajectory representation, the current MAV state, and the stereo camera baseline.
Since obstacle avoidance is inherently multimodal, multiple topologically distinct trajectories may be feasible for reaching the goal.
To preserve this diversity while mitigating mode collapse and suboptimal mode averaging, we adopt a primitive-anchored prediction representation \cite{yopo}.
Specifically, we introduce a set of learnable trajectory tokens $\mathbf{Z}_{\text{traj}} \in \mathbb{R}^{N\times D'}$, where each token $\mathbf{z}^i_{\mathrm{traj}}\in\mathbb{R}^{D'}$ is associated with a primitive anchor, as illustrated in Fig.~\ref{Fig: NetworkArchitecture}{\color{blue}{(d)}}, and is specialized to predict the terminal state of the corresponding topological trajectory:
\begin{equation}
\begin{aligned}
&{{\bf{Z}}_{{\rm{traj}}}} = \left[ { \cdots ,{\bf{z}}_{{\rm{traj}}}^i, \cdots } \right],i = 1, \ldots ,N\\
&{\bf{z}}_{{\rm{traj}}}^i \longleftrightarrow \left\{ {{{\bf{s}}_i}(T),{s_i}} \right\}
\end{aligned}
\end{equation}
where
$\mathbf{s}_i(T)$ and $s_i$ denote the terminal state of the associated topological trajectory and its corresponding score, respectively.

Although the stereo-camera extrinsics $\mathbf{T}_{\text{ext}}$ are provided to the Scene Representor (\ref{Eq: Scene Representor}) to extract geometry-aware features, the resulting scene representation remains scale-normalized because the VFM is trained with normalized geometry to maintain a consistent data distribution and enhance generalization.
Consequently, the scene representation does not explicitly preserve metric scale.
The stereo baseline $B$, however, provides an explicit metric-scale cue for reliable three-dimensional spatial perception.
To incorporate this geometric prior into the planning process, the baseline is treated as an additional state variable and concatenated with the MAV state observations, including the current velocity $\mathbf{v}$, acceleration $\mathbf{a}$, and goal direction $\mathbf{g}$.
\begin{equation}
{{\bf{Z}}_{{\rm{state}}}} = {\rm{MLP}}\left( {{{\left[ {{{\bf{v}}^\top},{{\bf{a}}^\top},{{\bf{g}}^\top},B} \right]}^\top}} \right)
\label{Eq: state_token}
\end{equation}
where $\mathbf{Z}_{\rm{state}} \in \mathbb{R}^{D'}$ is the state token encoded through the multilayer perceptron (MLP) block $\rm{MLP}(\cdot)$.

The trajectory tokens and the state token are first concatenated and processed by a self-attention block.
This operation allows each trajectory token to model its correlations with the metric-scale cue provided by the stereo baseline and with the current MAV state.
\begin{equation}
\begin{aligned}
&{\bf{Z}}_{{\rm{dec}}}^s = {\rm{MHA}}({\bf{Q}}_{{\rm{dec}}}^s,{\bf{K}}_{{\rm{dec}}}^s,{\bf{V}}_{{\rm{dec}}}^s),{{\bf{Z}}_{{\rm{dec}}}} = \left[ {{{\bf{Z}}_{{\rm{traj}}}},{{\bf{Z}}_{{\rm{state}}}}} \right]\\
&{\bf{Q}}_{{\rm{dec}}}^s = {{\bf{Z}}_{{\rm{dec}}}}{\bf{W}}_{{\rm{dec}}}^Q,{\bf{K}}_{{\rm{dec}}}^s = {{\bf{Z}}_{{\rm{dec}}}}{\bf{W}}_{{\rm{dec}}}^K,{\bf{V}}_{{\rm{dec}}}^s = {{\bf{Z}}_{{\rm{dec}}}}{\bf{W}}_{{\rm{dec}}}^V
\end{aligned} \label{Eq: self-attn}
\end{equation}
where ${\bf{Z}}_{{\rm{dec}}}^s \in \mathbb{R}^{(N+1)\times D'}$ denotes the self-attended representation of the concatenated trajectory and state tokens, and $\mathrm{MHA}(\cdot)$ denotes the multi-head attention block. The query, key, and value sequences ${\bf{Q}}_{{\rm{dec}}}^s$, ${\bf{K}}_{{\rm{dec}}}^s$, and ${\bf{V}}_{{\rm{dec}}}^s$ are obtained by projecting ${\bf{Z}}_{{\rm{dec}}}$ with the learnable matrices ${\bf{W}}_{{\rm{dec}}}^Q, {\bf{W}}_{{\rm{dec}}}^K, {\bf{W}}_{{\rm{dec}}}^V \in \mathbb{R}^{D'\times D'}$, respectively.

The resulting token sequence ${\bf{Z}}_{{\rm{dec}}}^s$ is then projected into queries and passed to a cross-attention block, where each trajectory query learns latent correspondences with the scene representation and retrieves geometry-aware information relevant to its associated topological trajectory.
\begin{equation}
\begin{aligned}
&{\bf{Z}}_{{\rm{dec}}}^c = {\rm{MHA}}({\bf{Q}}_{{\rm{dec}}}^c,{{\bf{K}}_{{\rm{enc}}}},{{\bf{V}}_{{\rm{enc}}}})\\
&{\bf{Q}}_{{\rm{dec}}}^c = \left( {{{\bf{Z}}_{{\rm{dec}}}} + {\bf{Z}}_{{\rm{dec}}}^s} \right){\bf{W}}_{{\rm{Asym}}}^Q
\end{aligned} \label{Eq: cross-attn}
\end{equation}
where ${\bf{Z}}_{{\rm{dec}}}^c \in \mathbb{R}^{(N+1)\times D}$ denotes the scene-conditioned representation produced by cross-attention, $\mathbf{K}_{\rm{enc}}$ and $\mathbf{V}_{\rm{enc}}$ are the cached key and value tensors from the final Transformer layer of the Scene Representor in (\ref{Eq: Scene Representor}), and $\mathbf{W}_{\rm{Asym}}^Q \in \mathbb{R}^{D'\times D}$ is a learnable asymmetric projection matrix that maps the decoder tokens to the encoder embedding space.

Following the sequential self- and cross-attention operations defined in (\ref{Eq: self-attn}) and (\ref{Eq: cross-attn}), the residual connections and feed-forward network of a standard Transformer block produce the refined trajectory-token sequence $\mathbf{Z}_{\rm{traj}}^{+}$.
An MLP head then predicts a terminal state and its corresponding score for each trajectory token. The terminal states are subsequently mapped to the candidate trajectories illustrated in Fig.~\ref{Fig: NetworkArchitecture}{\color{blue}{(e)}} through (\ref{eq_1}) and (\ref{eq_2}).
\begin{equation}
\hat{i} = \operatorname*{arg\,min}_{i\in\{1,\ldots,N\}} s_i \label{Eq: traj_score}
\end{equation}
Overall,
the Trajectory Decoder progressively decodes the geometric scene representations provided by the Scene Representor into multiple feasible trajectory candidates for quadrotor flight, from which the trajectory with the lowest predicted score is selected according to (\ref{Eq: traj_score}) for execution.

\begin{remark} \label{rem: compact decoder}
    \textbf{Interleaved Attention:} \textit{Unlike prior works\cite{black2024pi0,physicalintelligence2025pi05} that rely exclusively on joint attention, Stereo-Planner employs an interleaved approach, in which each block contains a self-attention stage and a cross-attention stage.
    Self-attention (\ref{Eq: self-attn}) enables the trajectory tokens to interact with one another and incorporate the MAV-state information, whereas cross-attention (\ref{Eq: cross-attn}) allows them to attend to the keys and values provided by the Scene Representor.
    Compared with joint-attention architectures, this decoupled attention scheme introduces a more structured inductive bias by separating inter-trajectory reasoning from trajectory-to-scene grounding.}
\end{remark}

\section{Training Strategy}

\begin{figure}[!t]\centering
\includegraphics[width=\linewidth]{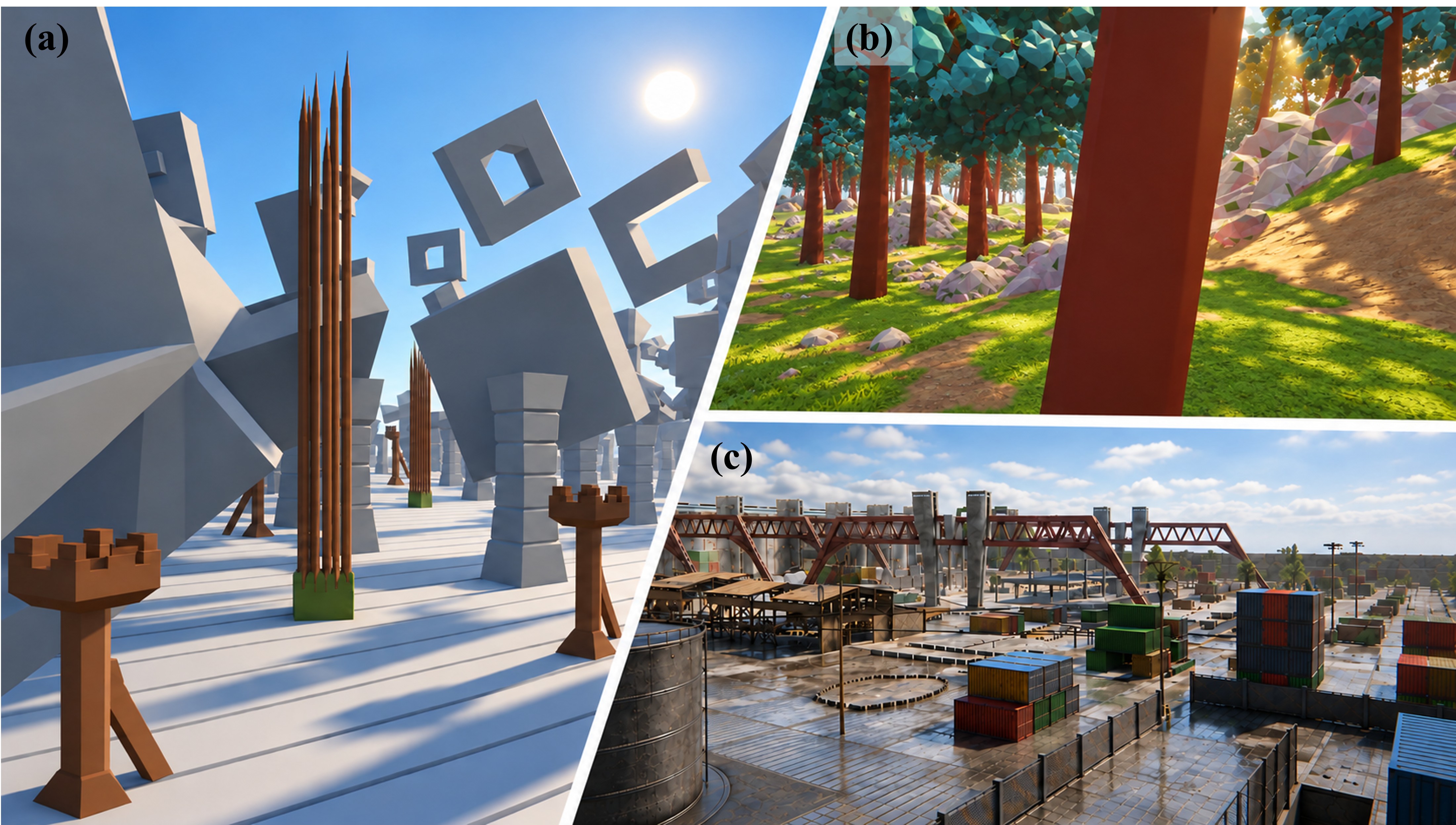}
\vspace{-1em}
\caption{Simulation scenarios. (a) Random obstacles. (b) Stylized forest. (c) Industrial.}\label{Fig: SimScene}
\vspace{-1em}
\end{figure}

To train and evaluate the proposed Stereo-Planner, we build a simulator based on the Unity game engine, which enables plug-and-play scene construction using assets from the Unity Asset Store\footnote{\url{https://assetstore.unity.com}}.
Three simulation scenarios are introduced, including a randomized environment (Fig.~\ref{Fig: SimScene}{\color{blue}(a)}) consisting of randomly generated obstacles for training data collection, as well as a stylized forest (Fig.~\ref{Fig: SimScene}{\color{blue}(b)}) and an industrial environment (Fig.~\ref{Fig: SimScene}{\color{blue}(c)}) for online evaluation in Section~\ref{Sec: Experimental Evaluations and Validations}.
By invoking the Unity rendering pipeline through C\# scripts, the simulator can synchronously render stereo RGB images and depth images from arbitrary poses.
Since 3D scene information is required during training and testing for ESDF generation and collision checking, we sample the mesh surfaces of all objects and terrain in each scene to generate the corresponding 3D point cloud.

\subsection{Data Collection and Augmentation in Simulation}\label{Sec: Data Collection and Augmentation in Simulation}

\subsubsection{Training Data Collection}\label{Sec: Training Data Collection}
During data collection, a point cloud map is provided for each scene, and numerous MAV poses are randomly sampled within the environment. To avoid collecting samples inside enclosed cavities or inaccessible regions, we leverage a pre-collected surface map to generate a large number of candidate sampling locations.
A sparse roadmap~\cite{kavraki1996prm} is then constructed by connecting collision-free neighboring samples, and only nodes belonging to the largest connected component are retained for data collection.
This efficiently removes isolated or unreachable regions while ensuring that all sampled poses are physically accessible. This offline pipeline eliminates expensive online simulation, significantly improving training efficiency while providing diverse scene coverage and facilitating data augmentation.

\subsubsection{Virtual Baseline Augmentation} \label{Sec: Virtual Baseline Augmentation}
\begin{figure}[!t]\centering
\includegraphics[width=\linewidth]{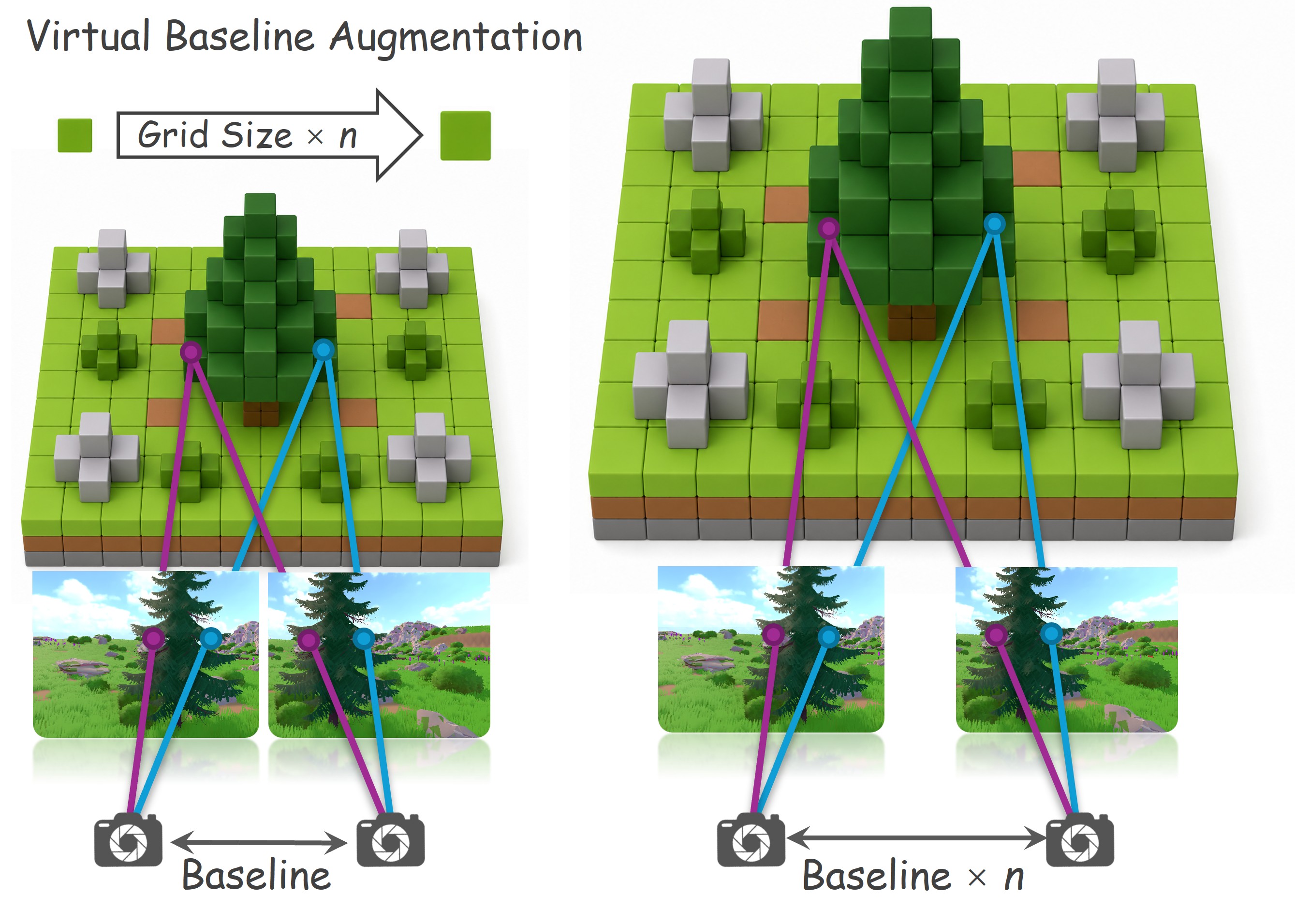}
\vspace{-1.2em}
\caption{Illustration of the proposed VBA strategy.
Without recomputing the ESDF, virtual scene scaling is achieved by modifying only the ESDF indexing and queried distances, allowing each stereo image pair to be associated with arbitrary virtual scene scales and corresponding stereo baselines for efficient data augmentation.}\label{Fig: VirtualBaselineAug}
\vspace{-1em}
\end{figure}

Considering that vision grounding foundation models are typically pretrained with normalized depth supervision rather than metric depth supervision to provide a more consistent learning target across diverse datasets, the geometric features extracted by the DA3-powered Scene Representor inherently exhibit scale ambiguity.
As described in Section~\ref{Sec: Trajectory Decoder}, the stereo baseline is explicitly provided to the Trajectory Decoder as a metric-scale cue, enabling it to infer the relationship between the scale-invariant geometric representations extracted by the Scene Representor and metric-scale motion planning.
To facilitate this learning process, we introduce a Virtual Baseline Augmentation (VBA) strategy.
For a stereo camera, the depth $Z$ associated with a pair of corresponding pixels can be calculated as:
\begin{equation}
Z = \frac{{fB}}{d} \label{Eq: Depth}
\end{equation}
where $f$ and $d$ denote the focal length and disparity, respectively.
Therefore, as illustrated in Fig.~\ref{Fig: VirtualBaselineAug}, scaling the baseline to $n \times B$ is equivalent to uniformly scaling the environment with the MAV's current position ${\bf{p}}(0)$ as the scaling center.
This enables a dataset collected with a fixed baseline to be reused for arbitrary stereo baselines.

To avoid rebuilding the ESDF after environment scaling, we modify the ESDF indexing and reading scheme.
The voxel index of a 3D position ${\bf{p}} \in \mathbb{R}^3$ is computed as:
\begin{equation}
{\texttt{idx}}({\bf{p}}) = \left\lfloor {\frac{{{\bf{p}}(0) + ({\bf{p}} - {\bf{p}}(0))/n - {\bf{o}}}}{r}} \right\rfloor 
\end{equation}
where ${\bf{o}}\in \mathbb{R}^3$ is the map origin and $r$ is the voxel resolution.

Interpreting the ESDF resolution as $n\times r$ and scaling the nearest-obstacle distance to $n\times \text{ESDF}(\texttt{idx})$ allows the ESDF to be queried as if the environment were uniformly scaled.
As a result, a single precomputed ESDF can be reused across arbitrary scaling factors, allowing each training sample to be assigned a random scale.
By exposing the network to diverse virtual stereo baselines during training, the proposed strategy encourages the Trajectory Decoder to infer metric-scale scene geometry from stereo disparity rather than scene appearance.
\subsection{Differentiable Training}
To ensure trajectory feasibility, we directly optimize the predicted trajectories in an end-to-end manner by jointly minimizing the loss function $\mathcal{L}$, which is designed as a linear combination of the safety cost $J_o$, the smoothness cost $J_s$, the goal-reaching cost $J_g$, and the trajectory performance estimation loss ${J_c}$.
\begin{equation}
\mathcal{L} = {\lambda _o}{J_o} + {\lambda _s}{J_s} + {\lambda _g}{J_g} + {\lambda _c}{J_c}
\end{equation}
where $\lambda_o$, $\lambda_s$, $\lambda _g$, and $\lambda _c$ are coefficients that balance the loss terms.

The safety cost is formulated as:
\begin{equation}
{J_o} = \int_0^T {F\left(n\,{\rm{ESDF}}\left( {\texttt{idx}}({\bf{p}}(t)) \right)\right)} dt\label{Eq: safety cost}
\end{equation}
where $F(\cdot)$ is an exponential potential function that suppresses the influence of distant obstacles while strongly penalizing near-collision states.

The smoothness cost is defined as the integral of the squared norm of a higher-order derivative of the trajectory.
\begin{equation}
{J_s} = \int_0^T {{{\left\| {{\dddot{\mathbf{p}}}(t)} \right\|}^2}} dt \label{Eq: smoothness}
\end{equation}
where ${\dddot{\mathbf{p}}}$ denotes the jerk, i.e., the third-order derivative of position.

The goal-reaching cost encourages the terminal state of the trajectory to progress toward the desired local goal $\mathbf{g}$.
\begin{equation}
{J_g} = \left\| {{\bf{p}}(T) - {\bf{g}}}  \right\|^2\label{Eq: goal cost}
\end{equation}

The trajectory performance estimation loss is defined as the mean squared error between the estimated trajectory score $\hat s$ and the corresponding ground-truth cost $s$:
\begin{equation}
{J_c} = \left( \hat s - s \right)^2\label{Eq: score cost}
\end{equation}
where $s = {\lambda _o}{J_o} + {\lambda _s}{J_s} + {\lambda _g}{J_g}$.
The cost functions defined in (\ref{Eq: safety cost})–(\ref{Eq: score cost}) are fully differentiable, allowing their gradients to be backpropagated directly to the network predictions for end-to-end optimization.

\section{Experimental Setup}\label{Sec: Experimental Setup}
In the experiments, the embedding dimension of the Scene Representor in (\ref{Eq: Scene Representor}) is set to $D=384$, and the Scene Representor is initialized with the pretrained weights of DA3-Small from DA3~\cite{depthanything3}.
For the Trajectory Decoder, we adopt a lower token dimension of $D'=192$, which is mapped to the encoder embedding space via the learnable asymmetric projection matrix $\mathbf{W}_\text{Asym}^Q$ introduced in (\ref{Eq: cross-attn}).
We use the Adam optimizer with a learning rate of $1.5\times10^{-4}$.
The resolution of the input stereo images is set to $H\times W= 504\times 280$.
All experiments reported in Section~\ref{Sec: Experimental Evaluations and Validations} are conducted on a computer equipped with an Intel Core i9-13900KF and an NVIDIA GeForce RTX 4080.
For a fair comparison, all benchmark methods are retrained for 50 epochs using their default configurations in the simulated random-obstacle environment shown in Fig.~\ref{Fig: SimScene}{\color{blue}(a)}. 

To demonstrate the effectiveness of the proposed method, we present detailed quantitative analyses comparing Stereo-Planner with SOTA vision-based quadrotor motion planning methods, including YOPO V2\cite{yopov2} and Ego-Planner V2\cite{doi:10.1126/scirobotics.abm5954}.
For the reference comparison, YOPO V2 and Ego-Planner V2 receive ground-truth depth images directly from the simulator.
Additional experiments replace these images with metric depth predicted from stereo observations by DA3\cite{depthanything3} to evaluate the effect of explicit depth recovery on planning performance and computational cost.
Details of the baseline methods are given as follows:   
\begin{itemize}
    \item \textit{YOPO V2}\cite{yopov2}: 
    an end-to-end depth-to-trajectory motion planning framework for quadrotors that employs depth images to mitigate the appearance gap between simulation and the real world, while jointly learning implicit scene perception, front-end path search, and back-end trajectory optimization within a single neural network.
    \item \textit{Ego-Planner V2}\cite{doi:10.1126/scirobotics.abm5954}: 
    a gradient-based motion planning framework for quadrotors that integrates local occupancy map construction and polynomial trajectory optimization, and performs ESDF-free optimization based on the MINCO\cite{minco} trajectory representation for agile and efficient flight.
\end{itemize}
An ablation study of Stereo-Planner is conducted under various configurations to investigate whether incorporating a vision grounding foundation model helps bridge the appearance gap, assess the effectiveness of the interleaved attention mechanism proposed in Section~\ref{Sec: Trajectory Decoder}, and determine whether the VBA strategy introduced in Section~\ref{Sec: Virtual Baseline Augmentation} enables the planner to recover absolute scale from the scale-normalized features provided by the Scene Representor.
\begin{itemize}
    \item \textit{Stereo-Planner}: The complete algorithm proposed in this paper achieves zero-shot sim-to-real end-to-end stereo-to-trajectory motion planning without relying on any depth sensors (e.g., depth cameras or LiDAR) or auxiliary depth estimation algorithms.
    \item \textit{Stereo-Planner w/o FM}:
    Stereo-Planner trained entirely from scratch without any foundation model initialization or parameter freezing.
    \item \textit{Stereo-Planner w. DINO}: Stereo-Planner with the Scene Representor introduced in Section~\ref{Sec: Scene Representor} initialized using the pretrained DINO V2 \cite{oquab2023dinov2} instead of the DA3 \cite{depthanything3} encoder, with its parameters kept frozen throughout training.
    \item \textit{Stereo-Planner w/o VBA}: Stereo-Planner trained without the VBA introduced in Section~\ref{Sec: Virtual Baseline Augmentation}.
    \item \textit{Stereo-Planner w. Joint-Attn.}: 
    Stereo-Planner with the interleaved attention described in Remark~\ref{rem: compact decoder} replaced by a joint-attention scheme, in which the trajectory queries attend to the concatenated trajectory, MAV-state, and scene keys and values within a single attention operation, following the design commonly adopted in SOTA VLA methods, such as $\pi_0$\cite{black2024pi0} and $\pi_{0.5}$\cite{physicalintelligence2025pi05}.
\end{itemize}

\section{Experimental Evaluations and Validations}\label{Sec: Experimental Evaluations and Validations}
For online evaluation, the quadrotor dynamics model runs in real time and provides poses to the Unity simulator introduced in Section~\ref{Sec: Data Collection and Augmentation in Simulation} through the ROS-TCP-Connector\footnote{\url{https://github.com/Unity-Technologies/ROS-TCP-Connector}}.
For a fair comparison, $100$ start-goal pairs are randomly generated in each of the forest and industrial environments shown in Fig.~\ref{Fig: SimScene}{\color{blue}(b)} and Fig.~\ref{Fig: SimScene}{\color{blue}(c)}, respectively, and all methods are evaluated on the same start-goal pairs at speeds of $4$, $6$, $8$, and $10\,\mathrm{m/s}$.
Five metrics are used to assess the performance of all methods listed in Section~\ref{Sec: Experimental Setup}.

\begin{itemize}
    \item \textit{Success Rate}: the percentage of successful runs over the total number of runs, where a run is considered successful if the MAV reaches the goal location while maintaining a minimum obstacle clearance of at least $0.1\,\mathrm{m}$ throughout the entire trajectory.

    \item \textit{Safety}: the average and minimum obstacle clearances over the entire trajectory.

    \item \textit{Velocity}: the average and maximum velocities along the trajectory.

    \item \textit{Smooth}: the integral of the squared jerk, corresponding to the smoothness cost defined in (\ref{Eq: smoothness}).

    \item \textit{Length}: the total length of the trajectory.
\end{itemize}

\begin{table*}[!t] \centering
\centering
\caption{Planning performance in the forest and industrial environments with an expected speed of $8 \rm{m/s}$. The best values among Stereo-Planner, YOPO V2, and Ego-Planner V2 are highlighted in \textbf{bold}.}
\renewcommand{\arraystretch}{1.2}
\label{tab: recall}
\label{tab: recall garden}\begin{minipage}{\textwidth}
\centering
\begin{minipage}[t]{0.49\linewidth}
\vspace{0pt}
\centering
\textbf{Forest}\par\smallskip
\scriptsize
\setlength{\tabcolsep}{2pt}
\begin{adjustbox}{max width=\linewidth}
\begin{threeparttable}
\begin{tabular}{l|| c c c c c}
\hline\hline
Method &\makecell{Success\\Rate} &\makecell{Safety (m)\\Mean/Min} &\makecell{Speed (m/s)\\Mean/Max} &Smooth &\makecell{Length (m)} \\\hline
Stereo-Planner &0.95 &\textbf{1.85}/0.59 &\textbf{8.12}/8.60 &\textbf{823.62} &\textbf{55.77} \\
YOPO V2 &\textbf{0.97} &1.83/\textbf{0.71} &7.99/8.23 &1114.11 &57.73 \\
Ego-Planner V2 &0.71 &1.46/0.24 &7.95/\textbf{8.82} &11771.72 &62.37 \\\hline
Stereo-Planner w. DINO &0.05 &2.06/0.10 &4.06/9.18 &2449.65 &62.37 \\
Stereo-Planner w/o FM &0.02 &2.60/0.02 &9.23/10.77 &27448.56 &139.13 \\
Stereo-Planner w/o VBA &0.84 &1.75/0.44 &7.99/8.50 &1156.82 &55.24 \\
Stereo-Planner w. Joint-Attn. &0.88 &1.69/0.42 &7.87/8.17 &904.59 &56.05 \\
\hline\hline
\end{tabular}
\end{threeparttable}
\end{adjustbox}
\end{minipage}\hfill\begin{minipage}[t]{0.49\linewidth}
\vspace{0pt}
\centering
\textbf{Industrial}\par\smallskip
\scriptsize
\setlength{\tabcolsep}{2pt}
\begin{adjustbox}{max width=\linewidth}
\begin{threeparttable}
\begin{tabular}{l|| c c c c c}
\hline\hline
Method &\makecell{Success\\Rate} &\makecell{Safety (m)\\Mean/Min} &\makecell{Speed (m/s)\\Mean/Max} &Smooth &\makecell{Length (m)} \\\hline
Stereo-Planner &\textbf{0.95} &\textbf{2.25}/\textbf{0.71} &\textbf{8.23}/8.91 &977.70 &\textbf{66.01} \\
YOPO V2 &0.93 &1.89/\textbf{0.71} &7.88/8.28 &\textbf{817.02} &67.35 \\
Ego-Planner V2 &0.79 &1.02/0.25 &8.06/\textbf{9.33} &8059.90 &70.77 \\\hline
Stereo-Planner w. DINO &0.62 &1.43/0.22 &7.87/8.12 &951.36 &66.42 \\
Stereo-Planner w/o FM &0.27 &7.81/0.30 &8.30/9.29 &5260.54 &153.44 \\
Stereo-Planner w/o VBA &0.88 &1.97/0.54 &8.04/8.79 &1856.74 &66.69 \\
Stereo-Planner w. Joint-Attn. &0.87 &1.75/0.50 &7.97/8.70 &1265.52 &66.38 \\
\hline\hline
\end{tabular}
\end{threeparttable}
\end{adjustbox}
\end{minipage}
\par\smallskip
{\scriptsize\raggedright
\textsuperscript{1} YOPO V2\cite{yopov2} and Ego-Planner V2\cite{doi:10.1126/scirobotics.abm5954} use ground-truth depth images provided by the simulator as input in both environments.\par}
\end{minipage}\vspace{-1em}
\end{table*}
\begin{figure}[!t]\centering
\includegraphics[width=\linewidth]{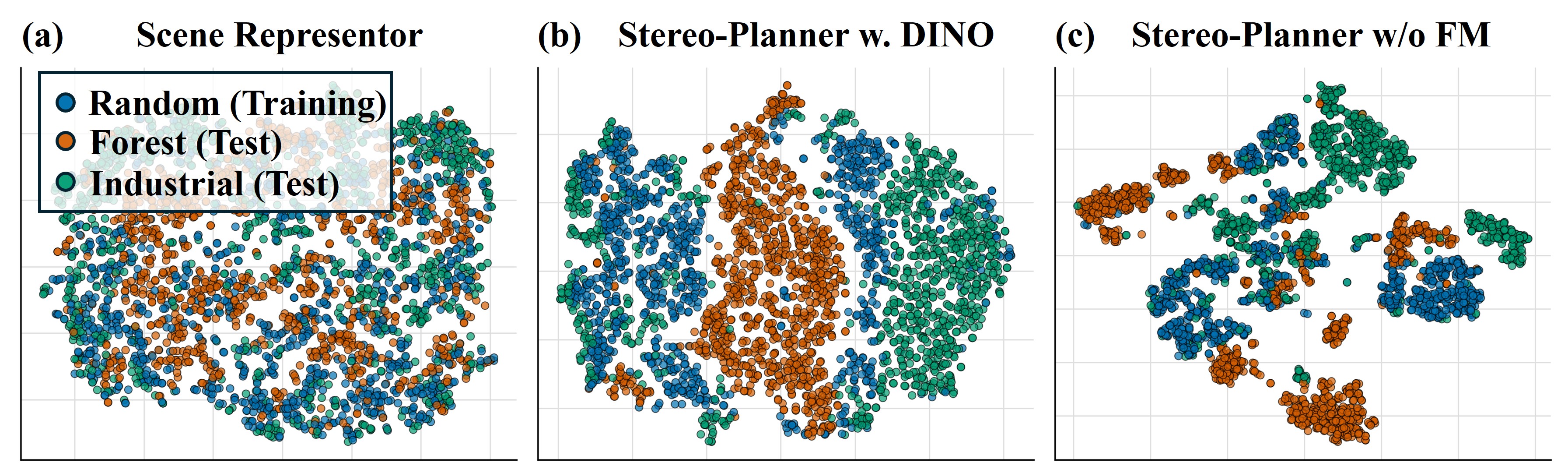}
\vspace{-1em}
\caption{t-SNE visualization of scene representations extracted from the random training environment and the unseen forest and industrial test environments.}\label{Fig: T-SNE-FM}
\vspace{-1.5em}
\end{figure}

\subsection{Ablation Study}
\subsubsection{Vision Grounding Foundation Model}
To evaluate the contribution of the vision grounding foundation model to zero-shot cross-scene generalization, we compare the proposed Stereo-Planner with \textit{Stereo-Planner w/o FM} and \textit{Stereo-Planner w. DINO}.
All three models are trained exclusively in the low-texture random environment (Fig.~\ref{Fig: SimScene}{\color{blue}(a)}) and evaluated without fine-tuning in the visually distinct forest (Fig.~\ref{Fig: SimScene}{\color{blue}(b)}) and industrial (Fig.~\ref{Fig: SimScene}{\color{blue}(c)}) environments.
Fig.~\ref{Fig: T-SNE-FM} visualizes the feature distributions of the three methods across the random training environment and the forest and industrial test environments, providing a qualitative explanation for their different generalization capabilities.
As shown in Fig.~\ref{Fig: T-SNE-FM}, the feature representations learned by \textit{Stereo-Planner w/o FM} differ substantially between the training environment and the two visually distinct test environments, suggesting limited generalization across appearance variations.
This scene-dependent organization suggests limited transferability of the representations learned from the low-texture training data, consistent with the low success rates of $2\%$ and $27\%$ in the forest and industrial environments reported in Table~\ref{tab: recall}.
With DINO V2 pretraining, Fig.~\ref{Fig: T-SNE-FM}\textcolor{blue}{(b)} shows greater overlap between the feature distributions of the training and industrial environments, whereas the forest features remain largely separated. 
This separation is attributed to the distinct visual appearance of the forest environment, whose textures and lighting differ substantially from those of both the random-obstacle training environment and the industrial test environment.
The resulting appearance gap limits generalization, consistent with the success rate of only $5\%$ achieved by \textit{Stereo-Planner w. DINO} in the forest environment.
In contrast, Fig.~\ref{Fig: T-SNE-FM}\textcolor{blue}{(a)} shows substantially greater overlap among the feature distributions of all three environments for the Scene Representor. This overlap indicates improved transferability of the learned representations, enabling the Trajectory Decoder trained in the random-obstacle environment to operate more reliably in both unseen test environments.
This observation supports the interpretation that DA3 pretraining helps preserve planning-relevant geometric cues across changes in visual appearance.
Accordingly, Stereo-Planner achieves a success rate of $95\%$ in both environments, improving upon \textit{Stereo-Planner w. DINO} by $90\%$ and $33\%$ and upon \textit{Stereo-Planner w/o FM} by $93\%$ and $68\%$ in the forest and industrial environments, respectively.
\begin{figure}[!t]\centering
\IfFileExists{Fig/TimeJointAttn.png}{\includegraphics[width=\linewidth]{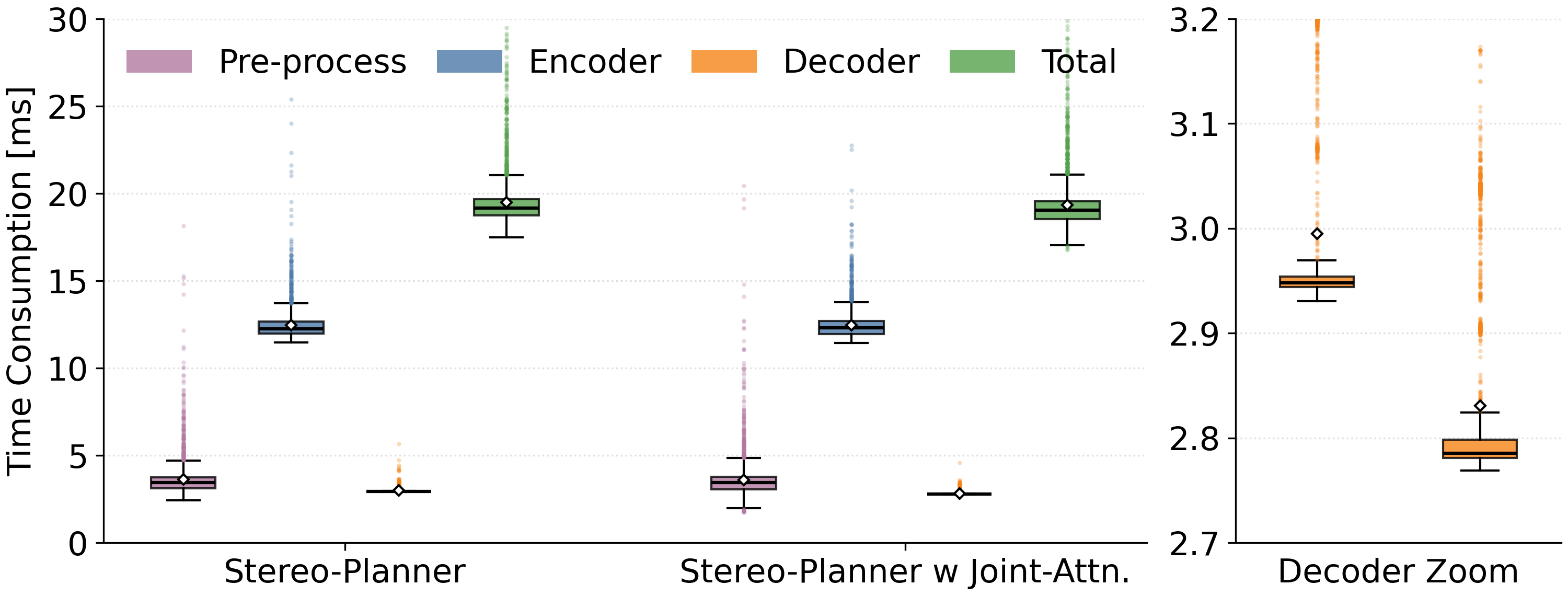}}{\fbox{\parbox[c][3cm][c]{0.94\linewidth}{\centering
Missing figure: \\ \texttt{Fig/TimeJointAttn.png}}}}
\vspace{-1em}
\caption{Time consumption of Stereo-Planner and \textit{Stereo-Planner w. Joint-Attn.}}\label{Fig: TimeJointAttn}
\vspace{-1em}
\end{figure}
\subsubsection{Interleaved Attention}
To evaluate the interleaved attention design described in Section~\ref{Sec: Trajectory Decoder}, we compare Stereo-Planner with \textit{Stereo-Planner w. Joint-Attn.}
As reported in Table~\ref{tab: recall}, interleaved attention increases the success rate from $88\%$ to $95\%$ in the forest environment and from $87\%$ to $95\%$ in the industrial environment. It also improves the average minimum obstacle clearance by $40.5\%$ and $42.0\%$ and reduces the smoothness cost by $9.0\%$ and $22.7\%$ in the forest and industrial environments, respectively.
These improvements are consistent with the structured inductive bias introduced by decoupling inter-trajectory reasoning from trajectory-to-scene grounding (Remark~\ref{rem: compact decoder}), and are achieved with only a small runtime overhead (as shown in Fig.~\ref{Fig: TimeJointAttn}).
To compare the attention operations, let $n=N+1$ and $\rho$ denote the numbers of trajectory-state tokens and scene tokens, respectively. 
Interleaved attention separates joint attention into self-attention (\ref{Eq: self-attn}) with complexity $\mathcal{O}(n^2D')$ and cross-attention (\ref{Eq: cross-attn}) with complexity $\mathcal{O}(n\rho D)$, which is lower than the $\mathcal{O}(n(n+\rho)D)$ complexity of joint attention.
The slight increase in inference latency can be attributed to the additional linear projection overhead introduced by the separate attention stages.
Since $n\ll\rho$, the computational cost of self-attention is small relative to that of scene retrieval. As shown in Fig.~\ref{Fig: TimeJointAttn}, both interleaved attention and joint attention require approximately $3\mathrm{ms}$ for trajectory decoding and $19\mathrm{ms}$ for the complete inference pipeline, with the Scene Representor accounting for most of the runtime.
Together with the planning results in Table~\ref{tab: recall}, these measurements indicate that interleaved attention improves planning reliability while introducing less than $0.2\mathrm{ms}$ of additional inference latency.

\subsubsection{Virtual Baseline Augmentation}
\begin{figure*}[!t]\centering
\includegraphics[width=\linewidth]{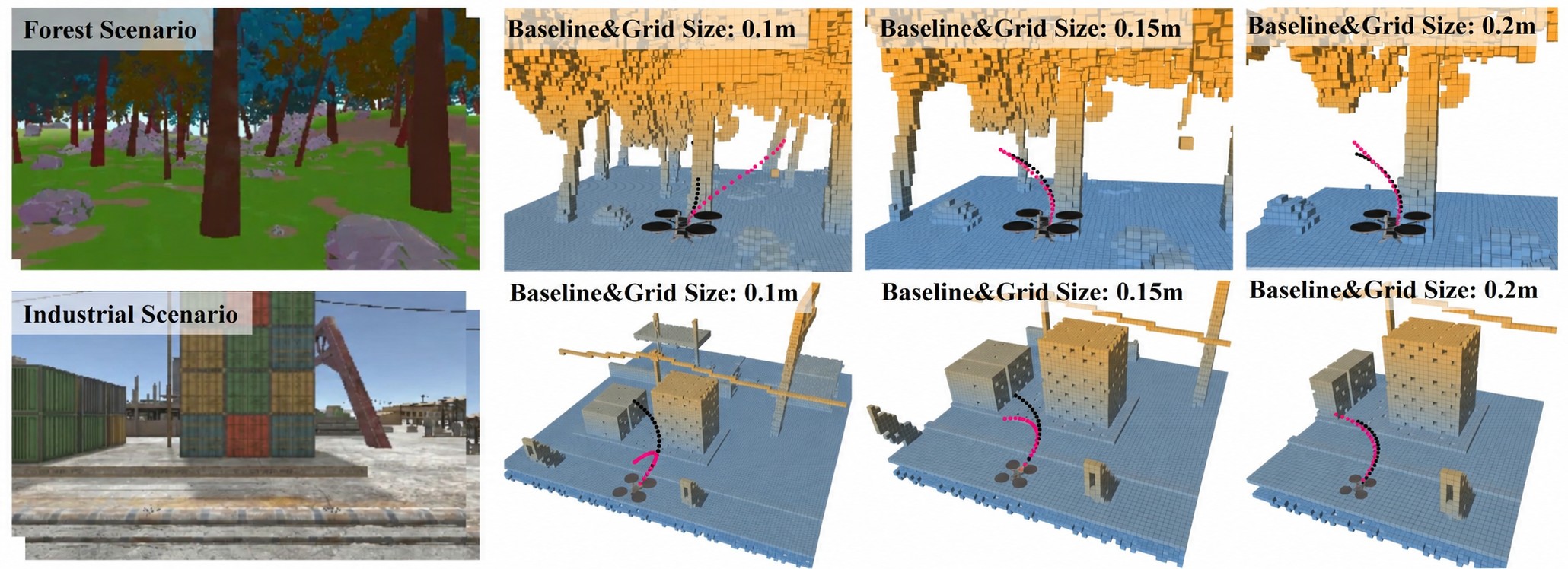}
\vspace{-1em}
\caption{Trajectory predictions under different virtual stereo baselines in forest (top) and industrial (bottom) scenarios. Within each scenario, the same stereo image pair is associated with baselines of $0.10$, $0.15$, and $0.20\,\mathrm{m}$, with the voxel grid size scaled accordingly to represent different metric scene scales. The visualized map covers a $10\times10\,\mathrm{m}$ area for all baselines. The trajectories illustrate the comparison between Stereo-Planner (red trajectories) and its variant without VBA (black trajectories).}\label{Fig: ExpVirtualBaselineAug}
\vspace{-1em}
\end{figure*}
To evaluate the contribution of VBA to metric-scale interpretation, we compare Stereo-Planner with \textit{Stereo-Planner w/o VBA}. In Fig.~\ref{Fig: ExpVirtualBaselineAug}, the same stereo image pair is associated with baselines of $0.10$, $0.15$, and $0.20\,\mathrm{m}$ in each scenario, with the environment scaled accordingly. Since the disparity remains unchanged, the metric depth varies proportionally with the baseline according to (\ref{Eq: Depth}). The planner is therefore required to account for the stereo baseline to distinguish between visually identical observations corresponding to different obstacle distances and free-space dimensions.
As shown in Fig.~\ref{Fig: ExpVirtualBaselineAug}, the trajectories predicted by \textit{Stereo-Planner w/o VBA} remain nearly unchanged across the three baseline settings. This behavior suggests that the model relies on the scene scale associated with the fixed $0.20\,\mathrm{m}$ training baseline, rather than adapting its predictions to the supplied baseline.
In contrast, the augmentation strategy introduced in Section~\ref{Sec: Virtual Baseline Augmentation} exposes Stereo-Planner to an explicit correspondence between baseline and metric scene scale, enabling it to adapt its trajectory to the actual obstacle layout.
In the forest example, Stereo-Planner adapts its avoidance trajectory to the stereo baseline, passing on different sides of the tree under different baseline settings, whereas \textit{Stereo-Planner w/o VBA} exhibits behavior consistent with an overestimation of the obstacle distance, resulting in collisions at smaller scales.
Similarly, in the industrial example with a $0.10\,\mathrm{m}$ baseline, the reduced free space makes the trajectory predicted by \textit{Stereo-Planner w/o VBA} collide with the container, while Stereo-Planner generates a braking trajectory that stops the vehicle before a collision occurs.

\begin{figure}[!t]
    \centering
    \subfigure[Forest]{\includegraphics[width=0.49\linewidth]{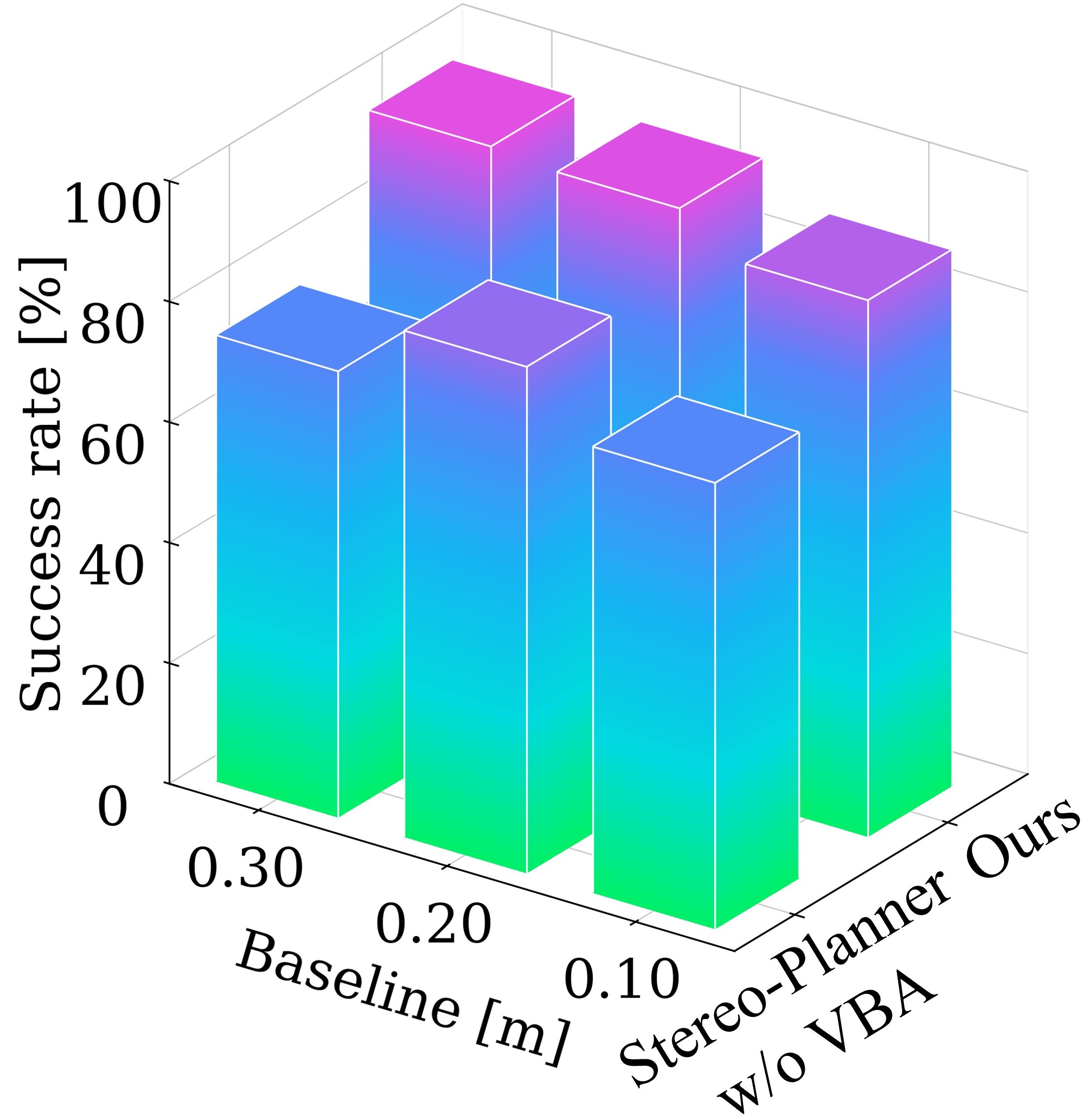}\label{Fig: SRVBAForest}}\hfill
    \subfigure[Industrial]{\includegraphics[width=0.49\linewidth]{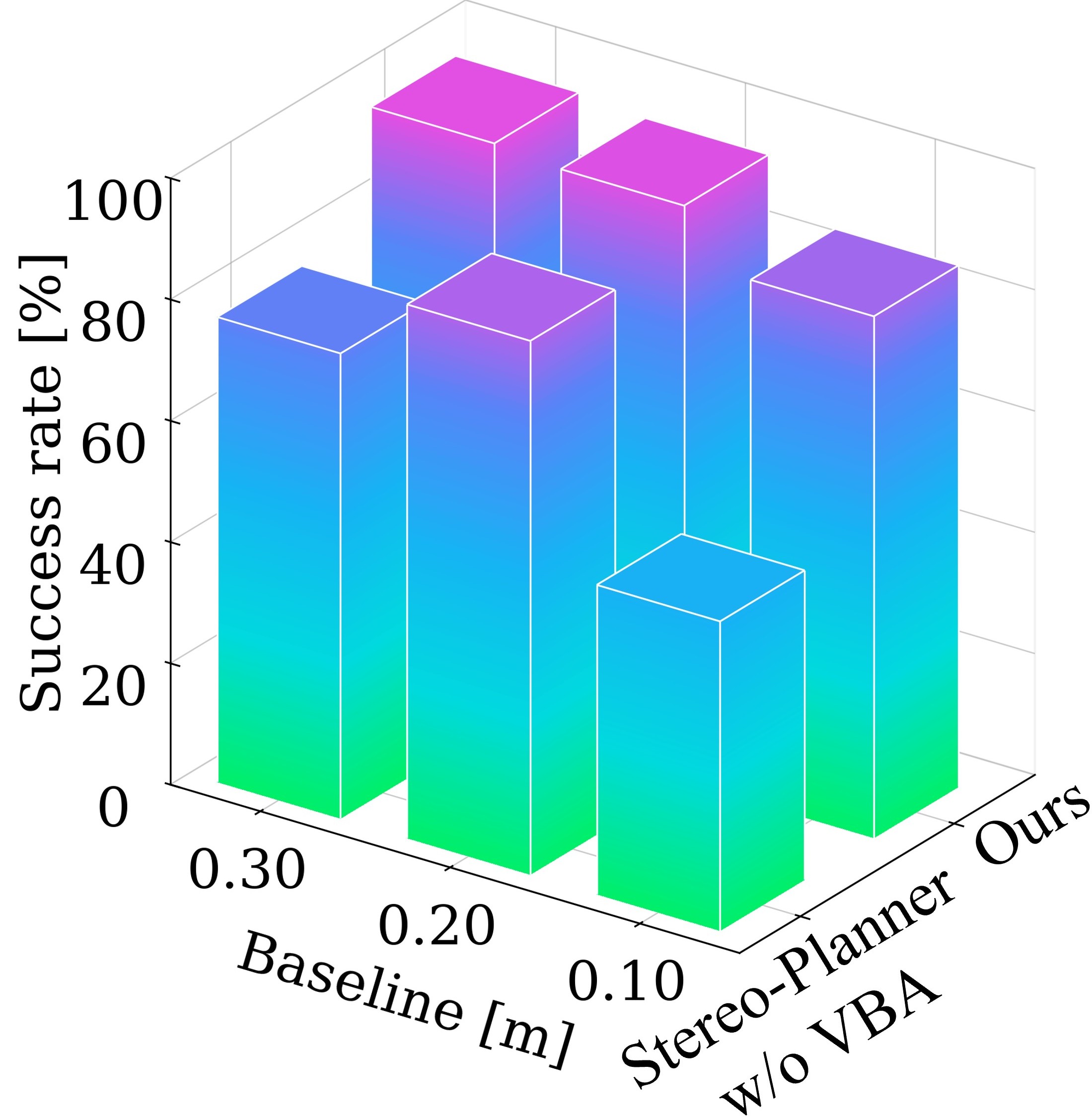}\label{Fig: SRVBAIndustrial}}
    \caption{Planning success rates with and without VBA across different stereo baselines in the forest and industrial environments.}
    \label{Fig: SRVBA}
    \vspace{-1.2em}
\end{figure}

We further evaluate planning success rates at stereo baselines of $0.30$, $0.20$, and $0.10\,\mathrm{m}$, as shown in Fig.~\ref{Fig: SRVBA}.
Under comparable imaging and correspondence conditions, a larger stereo baseline provides stronger geometric cues for scene representation, potentially improving the navigation success rate.
Nevertheless, \textit{Stereo-Planner w/o VBA} achieves lower success rates at both $0.30$ and $0.10\,\mathrm{m}$ than at its training baseline of $0.20\,\mathrm{m}$. 
This degradation is consistent with overfitting to the scene scale associated with the fixed training baseline ($0.20\,\mathrm{m}$), preventing the model from reliably interpreting metric geometry when the baseline changes.
By incorporating the VBA strategy introduced in Section~\ref{Sec: Virtual Baseline Augmentation}, Stereo-Planner learns to account for baseline-dependent metric scale and maintains consistently high success rates across the evaluated baselines.
The quantitative results in Table~\ref{tab: recall} are consistent with these observations. At a commanded speed of $8\,\mathrm{m/s}$, VBA increases the success rate from $84\%$ to $95\%$ in the forest environment and from $88\%$ to $95\%$ in the industrial environment.
Compared with \textit{Stereo-Planner w/o VBA}, the minimum obstacle clearances increase by $34.09\%$ and $31.48\%$, while the smoothness costs decrease by $28.80\%$ and $47.34\%$ in the forest and industrial environments, respectively. 
These results support the interpretation that VBA helps the planner use the baseline to resolve metric scale, thereby improving both collision avoidance and trajectory quality.

\subsection{Benchmark with Baseline Methods}

\begin{figure*}[!t]
    \centering
    \includegraphics[width=\linewidth]{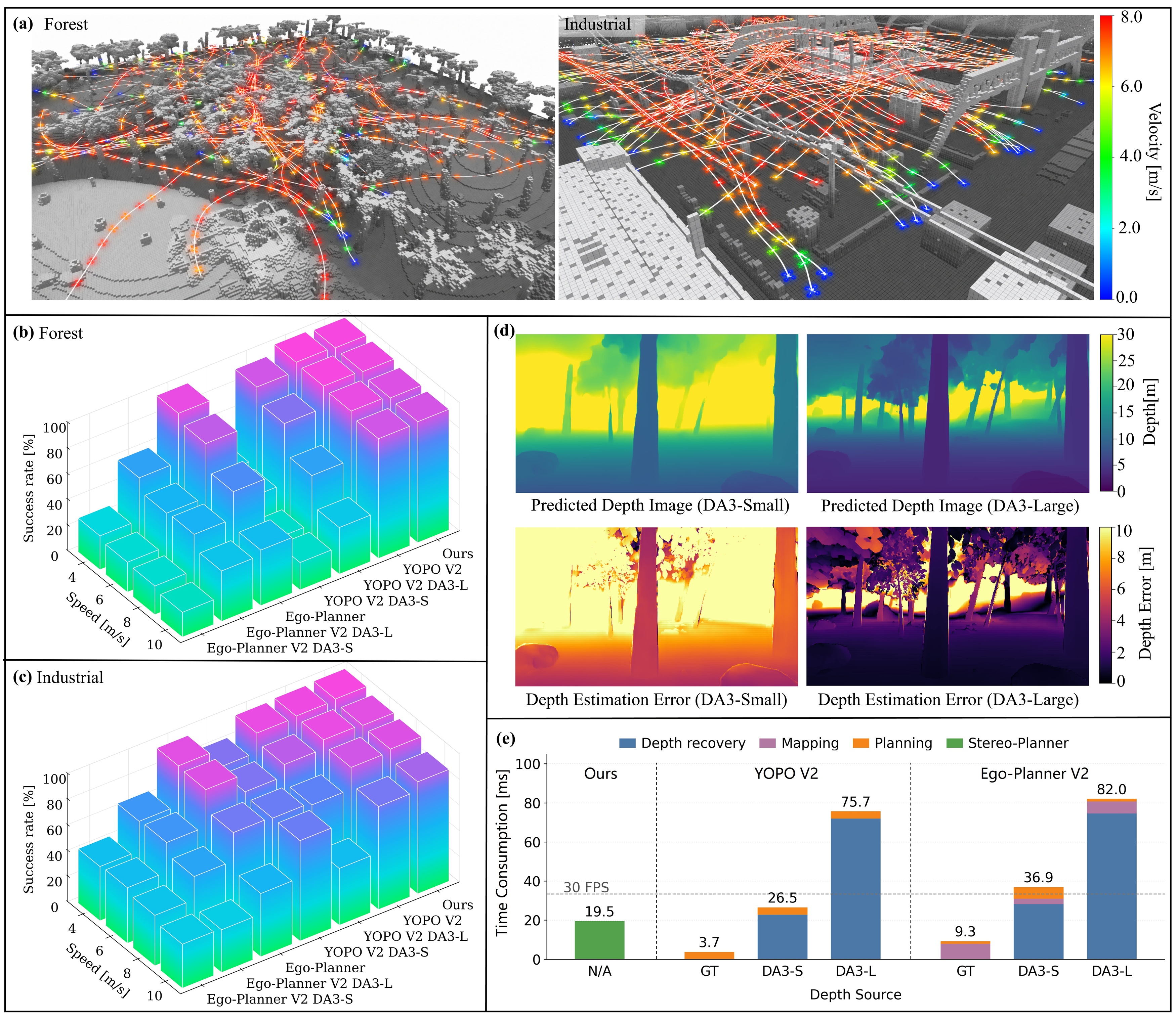}
    \vspace{-1.5em}
    \caption{Benchmark results of Stereo-Planner against baseline methods. (a) Executed flight trajectories of Stereo-Planner in the forest and industrial environments. (b) and (c) demonstrate the success rates at different commanded speeds in the forest and industrial environments, respectively. (d) Depth estimation comparison in the forest environment. Top row: predicted depth maps. Bottom row: absolute depth estimation errors relative to ground truth. (e) Computation time comparison of Stereo-Planner, YOPO V2, and Ego-Planner V2. GT denotes simulator-provided ground-truth depth; DA3-S and DA3-L denote depth predicted by DA3-Small and DA3-Large.}
    \label{Fig: Trajsim}
    \vspace{-1em}
\end{figure*}

Fig.~\ref{Fig: Trajsim}{\color{blue}(a)} illustrates the agile flight trajectories executed by the quadrotor using Stereo-Planner in the forest and industrial environments. The planner infers trajectories directly from stereo-RGB observations, which are then converted into desired attitude and thrust commands for tracking. 
Stereo-Planner achieves performance comparable to YOPO V2~\cite{yopov2}, which uses simulator-provided ground-truth depth, while improving the success rate over Ego-Planner V2~\cite{doi:10.1126/scirobotics.abm5954} by $24\%$ and $16\%$ in the forest and industrial environments, respectively.
Fig.~\ref{Fig: Trajsim}{\color{blue}(b)} and Fig.~\ref{Fig: Trajsim}{\color{blue}(c)} further compare the success rates of different methods across commanded speeds. The results suggest that Ego-Planner V2 becomes less reliable at speeds above $6\mathrm{m/s}$, whereas Stereo-Planner and YOPO V2 retain comparatively high success rates.
This trend is attributable to the map-based planning pipeline of Ego-Planner V2, in which trajectory generation relies on the map update process.
Probabilistic occupancy updates require the accumulation of multiple hit or miss observations to mitigate sensing noise, thereby introducing latency in obstacle detection. The resulting delay constrains the temporal margin for collision avoidance, especially during high-speed flight.
In contrast, Stereo-Planner and YOPO V2 directly infer trajectories from stereo RGB and depth images, respectively, maintaining strong planning performance even at higher flight speeds.

In Fig.~\ref{Fig: Trajsim}{\color{blue}(b)} and Fig.~\ref{Fig: Trajsim}{\color{blue}(c)}, the sensitivity of YOPO V2 and Ego-Planner V2 to depth estimation quality is also evaluated, reflecting the effects of depth prediction errors and noise on planning performance.
The depth inputs to YOPO V2 and Ego-Planner V2 are replaced with metric depth estimated from stereo observations using the DA3-Small and DA3-Large variants of DA3\cite{depthanything3}.
The resulting configurations are denoted by the suffixes DA3-S and DA3-L, respectively.
The inaccurate metric depth estimates produced by DA3-Small, as illustrated in Fig.~\ref{Fig: Trajsim}{\color{blue}(d)}, compromise the planning performance of both Ego-Planner V2 DA3-S and YOPO V2 DA3-S, even at a low flight speed of $4\mathrm{m/s}$.
DA3-Large provides more accurate metric depth estimates than DA3-Small, thereby improving the planning performance of both Ego-Planner V2 and YOPO V2. However, its $0.35$ billion parameters incur substantial computational overhead, resulting in total latencies of $82.0\,\mathrm{ms}$ and $75.7\,\mathrm{ms}$ for Ego-Planner V2 DA3-L and YOPO V2 DA3-L, respectively (Fig.~\ref{Fig: Trajsim}{\color{blue}(e)}). The resulting delays contribute to a marked decline in planning performance as flight speed increases.
By combining the geometry-aware DA3-Small encoder with a lightweight Trajectory Decoder, Stereo-Planner enables efficient end-to-end vision-to-trajectory inference while maintaining strong planning performance. Across commanded speeds from $4$ to $10\mathrm{m/s}$, it improves the success rate over YOPO V2 DA3-L and Ego-Planner V2 DA3-L by $6\%$–$58\%$ and $21\%$–$46\%$, respectively, in the forest environment, and by $2\%$–$43\%$ and $13\%$–$31\%$, respectively, in the industrial environment. Its end-to-end inference time is $19.5\,\mathrm{ms}$, corresponding to latency reductions of $74.2\%$ and $76.2\%$ relative to YOPO V2 DA3-L and Ego-Planner V2 DA3-L, respectively (Fig.~\ref{Fig: Trajsim}{\color{blue}(e)}).

\begin{figure}[!t]
\centering
\includegraphics[width=\linewidth]{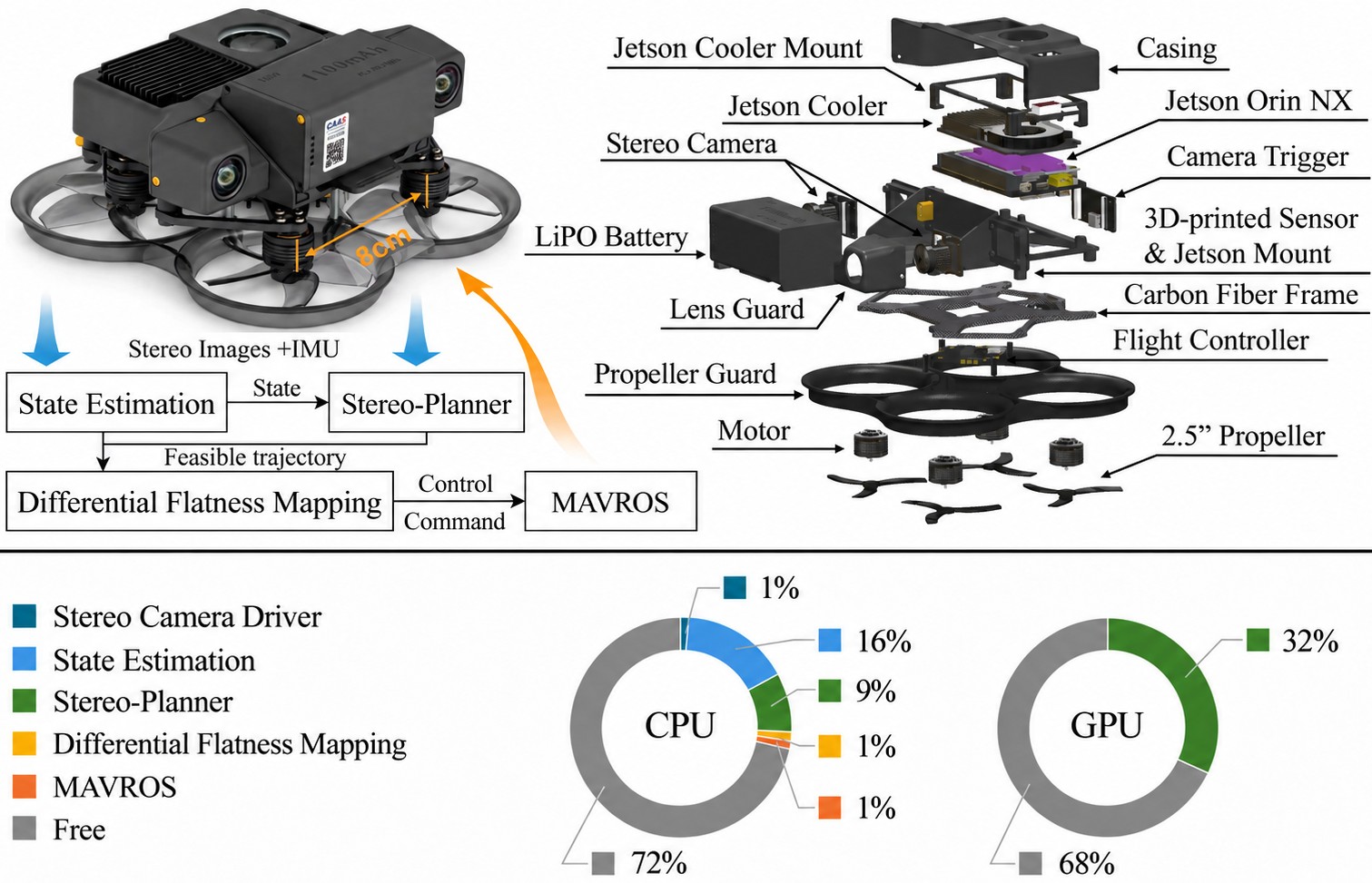}
\vspace{-1em}
\caption{Overview of Stereo-MAV. Top left: the assembled quadrotor with an $8\,\mathrm{cm}$ spacing between the front motor axes. The onboard software pipeline integrating visual-inertial state estimation, Stereo-Planner, and differential flatness mapping. Top right: an exploded view of the hardware components. Bottom: CPU and GPU utilization of the onboard modules on the NVIDIA Orin NX.}
\label{Fig: Platform}
\vspace{-1em}
\end{figure}
\section{Zero-shot Flight in the Real World}
To evaluate its applicability to real-world autonomous flight, Stereo-Planner is deployed on a compact quadrotor, termed Stereo-MAV, equipped with an NVIDIA Orin NX onboard computer, a stereo RGB camera, a flight control unit, and a propulsion system.
These components are integrated into a compact airframe, as depicted in Fig.~\ref{Fig: Platform}, which has a takeoff weight of $420\,\mathrm{g}$ and a wheelbase of only $112\,\mathrm{mm}$.
During autonomous flight, the software pipeline consists of three main components: state estimation, trajectory planning, and differential flatness mapping, which are integrated and coordinated through ROS2\cite{ros2}.
The state estimator fuses stereo observations with measurements from the flight controller's built-in inertial measurement unit to provide real-time vehicle state feedback for the state token in (\ref{Eq: state_token}).
Conditioned on the stereo images and state feedback, Stereo-Planner generates candidate trajectories and selects a trajectory for execution. The differential flatness mapping module converts the desired trajectory into attitude and thrust commands, which are sent to the flight control unit through MAVROS\footnote{\url{https://github.com/mavlink/mavros}}.

\begin{figure*}[!t]
\centering
\includegraphics[width=\linewidth]{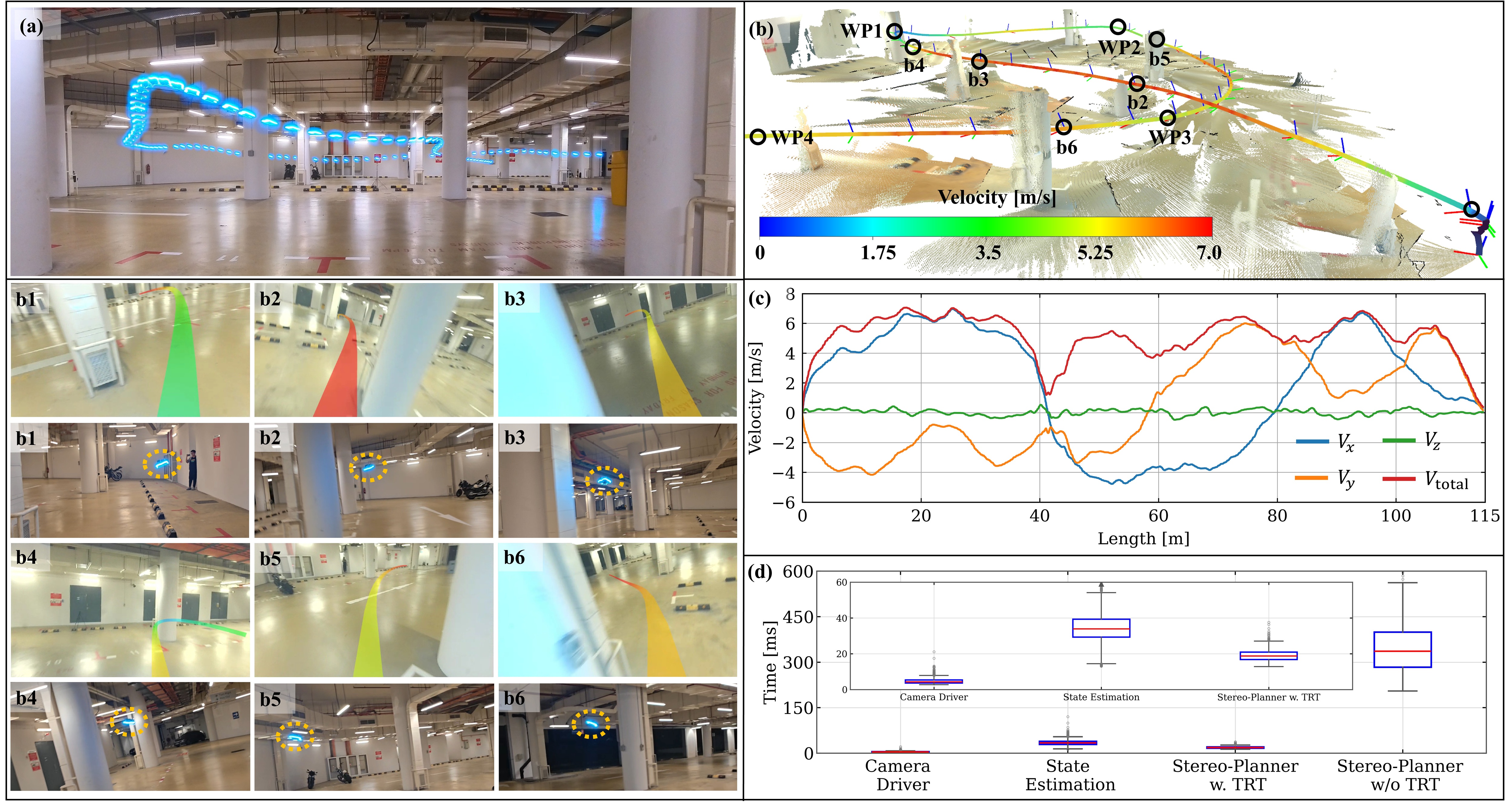}
\vspace{-1em}
\caption{Autonomous flight in a carpark using Stereo-Planner. (a) Overview of the executed flight paths.
(b) Executed flight trajectory color-coded by flight speed, passing through waypoints WP1–WP4. (c) Speed profiles along each axis during the real-world flight test, with a maximum speed exceeding $7 \mathrm{m/s}$. (d) Onboard processing time on the NVIDIA Orin NX deployed on the Stereo-MAV.}
\label{Fig: FlightCarpark}
\end{figure*}
\subsection{Vision-based Agile Flight in Carpark}
We evaluate Stereo-Planner in a previously unseen carpark, where the quadrotor navigates from an initial position to waypoints in sequence. The planner is trained using simulated data and deployed without fine-tuning on data from the test environment or any other real-world environment.
Fig.~\ref{Fig: FlightCarpark}\textcolor{blue}{(a)} shows the agile and aggressive flight trajectory executed by Stereo-MAV.
The snapshots in Fig.~\ref{Fig: FlightCarpark}\textcolor{blue}{(b1)-(b6)}, captured by the onboard camera and a third-person-view camera tracking the quadrotor, correspond to the marked positions along the flight trajectory shown in Fig.~\ref{Fig: FlightCarpark}\textcolor{blue}{(b)}.
The corresponding onboard image sequence shows that Stereo-Planner continues to generate executable collision-avoidance trajectories despite visible image blur induced by rapid motion and illumination variations caused by artificial lighting.
These observations demonstrate the feasibility of deploying the simulation-trained Stereo-Planner in real-world flight and its ability to navigate the carpark with agility, with the quadrotor reaching a maximum speed of approximately $7.12\mathrm{m/s}$ (Fig.~\ref{Fig: FlightCarpark}\textcolor{blue}{(c)}).

Fig.~\ref{Fig: FlightCarpark}\textcolor{blue}{(d)} reports the processing time of Stereo-Planner on the NVIDIA Orin NX. To meet the real-time requirements of onboard deployment, we use TensorRT with FP16 inference and Tensor Core acceleration, reducing the average network inference time from $336.16\mathrm{ms}$ to $18.99\,\mathrm{ms}$.
Differential flatness mapping and MAVROS communication are omitted from the timing plot because their respective processing times are below $1\mathrm{ms}$. During flight, state estimation is pinned to two CPU cores using the Linux \texttt{taskset} command and runs concurrently with Stereo-Planner. Including image preprocessing and differential flatness mapping, the mean vision-to-action processing time is $24.27\mathrm{ms}$, corresponding to an average processing capacity of approximately $41.2\mathrm{Hz}$ and supporting trajectory updates at $30\mathrm{Hz}$.
Fig.~\ref{Fig: Platform} summarizes the average CPU and GPU utilization of the onboard modules. The CPU handles image preprocessing, state estimation, and trajectory-to-action mapping, while the GPU performs network inference. This division of computation enables the deployment of a VFM-based trajectory planner on a compact autonomous quadrotor, with Stereo-Planner using an average of $8.60\%$ of the CPU capacity and $31.75\%$ of the GPU capacity.

\begin{figure}[!t]
    \centering
    \includegraphics[width=0.9\linewidth]{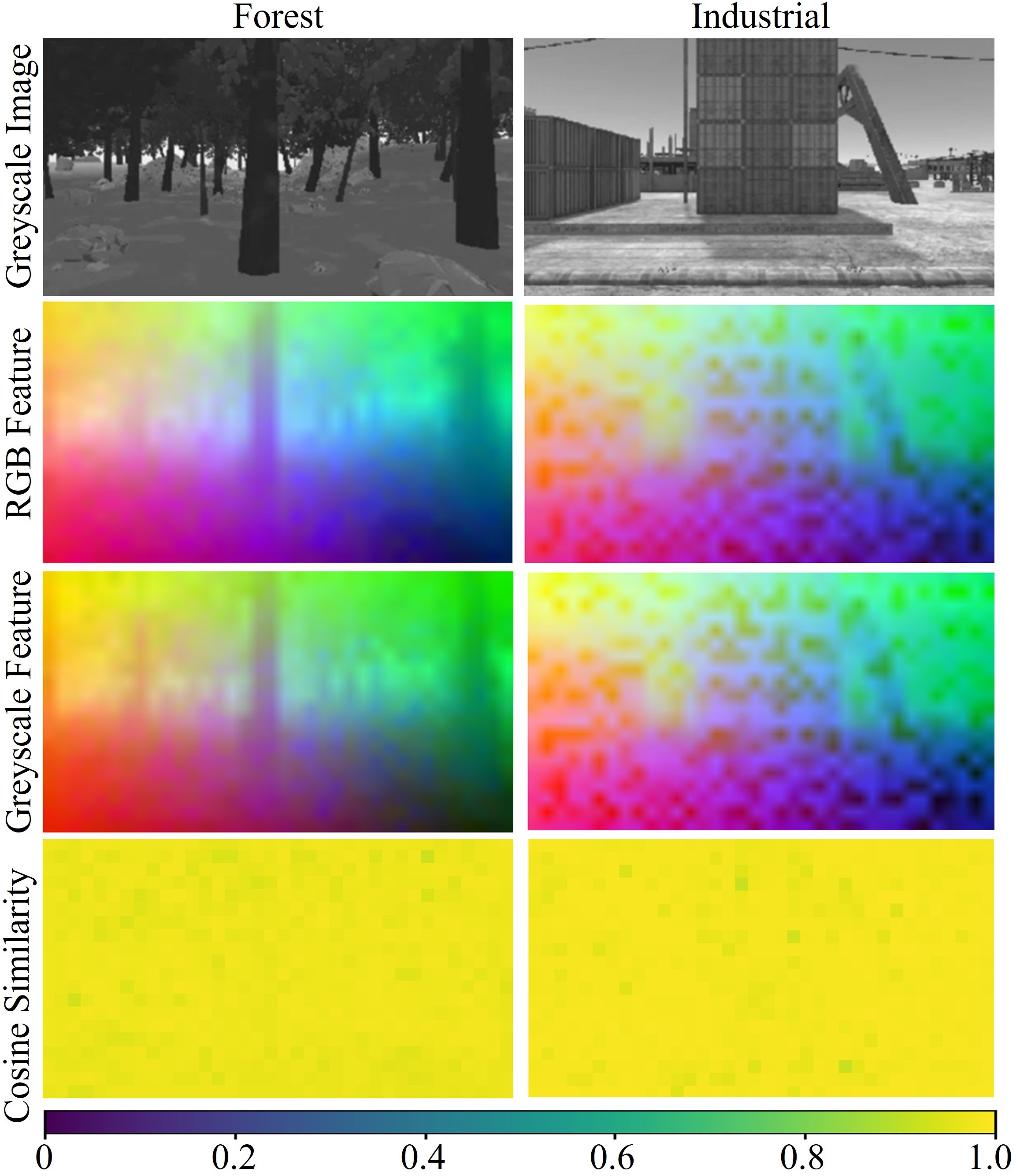}
    \caption{Token-wise feature visualizations (the two middle rows) of the Scene Representor for the RGB images in Fig.~\ref{Fig: ExpVirtualBaselineAug} and their grayscale counterparts (top row) in the forest and industrial simulation environments, along with cosine similarity maps between corresponding scene tokens (bottom row).}
    \label{Fig: RGBGrey}
    \vspace{-1em}
\end{figure}
\subsection{Zero-Shot Flight with Grayscale Images}
\begin{figure*}[!t]
    \centering
    \includegraphics[width=\linewidth]{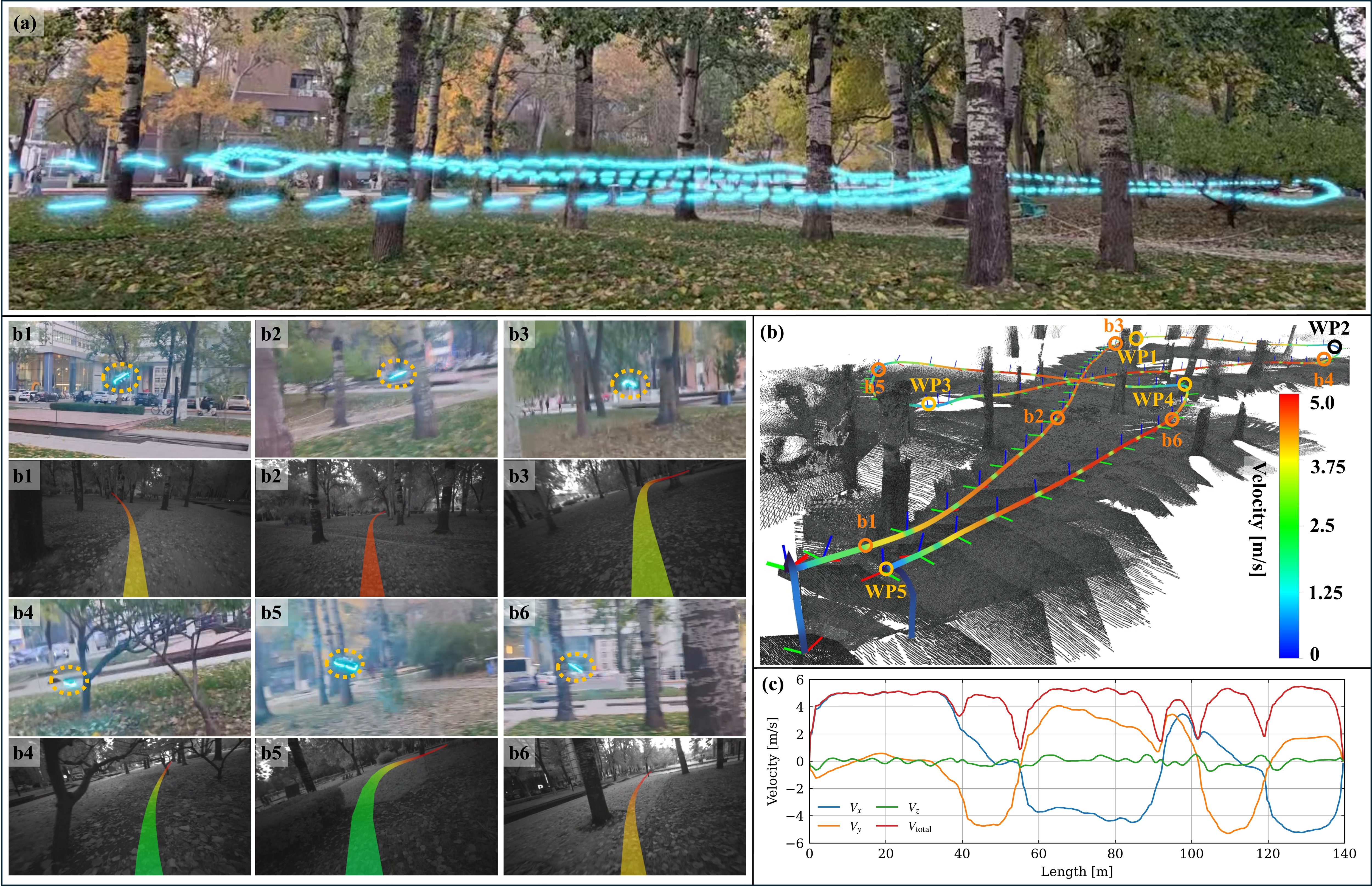}
    \vspace{-1.5em}
    \caption{Zero-shot autonomous flight in a forest environment using grayscale stereo images with Stereo-Planner. (a) Overview of the executed flight paths. (b) Executed flight trajectory color-coded by flight speed, passing through waypoints WP1–WP5. (c) Speed profiles along each axis during the real-world flight test, reaching a maximum speed of $5.5\,\mathrm{m/s}$.}
    \label{Fig: GreyFlight}
    \vspace{-1em}
\end{figure*}

Considering that some quadrotor platforms employ off-the-shelf stereo cameras, such as the Intel RealSense D400 series, whose stereo image pairs are monochrome rather than RGB, we further evaluate the zero-shot generalization capability of Stereo-Planner to grayscale stereo observations.
To accommodate the three-channel RGB input expected by the pretrained ViT\cite{ViT}-based Scene Representor, we replicate each single-channel grayscale image across three channels to construct a pseudo-RGB input.
This simple yet effective adaptation strategy enables Stereo-Planner to directly process grayscale stereo observations without architectural modifications or additional fine-tuning.
Owing to the pretrained vision grounding foundation model, DA3\cite{depthanything3}, as illustrated in Fig.~\ref{Fig: RGBGrey}, the Scene Representor extracts geometry-aware features that are less sensitive to appearance differences between RGB and grayscale images. To further evaluate whether this representation consistency translates into planning performance, we assess Stereo-Planner using grayscale stereo observations in both the forest and industrial simulation environments. Stereo-Planner achieves success rates of $94\%$ and $95\%$, respectively, demonstrating performance comparable to that obtained with RGB observations ($95\%$ in both environments as shown in Table~\ref{tab: recall}).
To evaluate the practical applicability of Stereo-Planner to MAV platforms equipped with grayscale stereo cameras, we mount an Intel RealSense D455 on the MAV and disable its depth and RGB streams, retaining only the stereo infrared image streams.
As shown in Fig.~\ref{Fig: GreyFlight}, Stereo-Planner achieves zero-shot, end-to-end grayscale-to-trajectory autonomous flight in a cluttered forest environment without any fine-tuning, reaching a maximum flight speed of $5.5\,\mathrm{m/s}$.

\section{Conclusion}
In this article, we have developed Stereo-Planner, an end-to-end vision-to-trajectory framework for zero-shot sim-to-real autonomous flight using stereo RGB observations. Stereo-Planner explores the use of a vision grounding foundation model to enhance the geometry-aware representation capability of the planning encoder, enabling domain-consistent scene representations across environments with diverse visual appearances.
We have also designed a compact Trajectory Decoder with interleaved attention to efficiently decode these geometric representations into feasible quadrotor trajectories. Benefiting from these designs, Stereo-Planner exhibits strong cross-scene generalization.
Although trained only in randomly generated low-texture environments, it achieves high-performance planning in visually and structurally distinct forest and industrial scenarios. Furthermore, through integrated software-hardware co-design, we have developed a complete open-source autonomous MAV system and investigated the real-time deployment of the vision-grounding-enhanced planner on onboard edge-computing platforms.
In Appendix~\ref{app: Distill}, we further present a self-supervised pruning and distillation pipeline for the Scene Representor that reduces its computational footprint while preserving the geometry-aware representations provided by the vision grounding foundation model pretrained on large-scale datasets.
The resulting system enables fully onboard autonomous flight in previously unseen real-world environments at speeds approaching $8\,\mathrm{m/s}$, demonstrating the effectiveness, generalization capability, and practical deployability of Stereo-Planner.

\appendix[Can the Scene Representor Be Distilled?]\label{app: Distill}
\begin{figure}[!t]
    \centering
    \includegraphics[width=\linewidth]{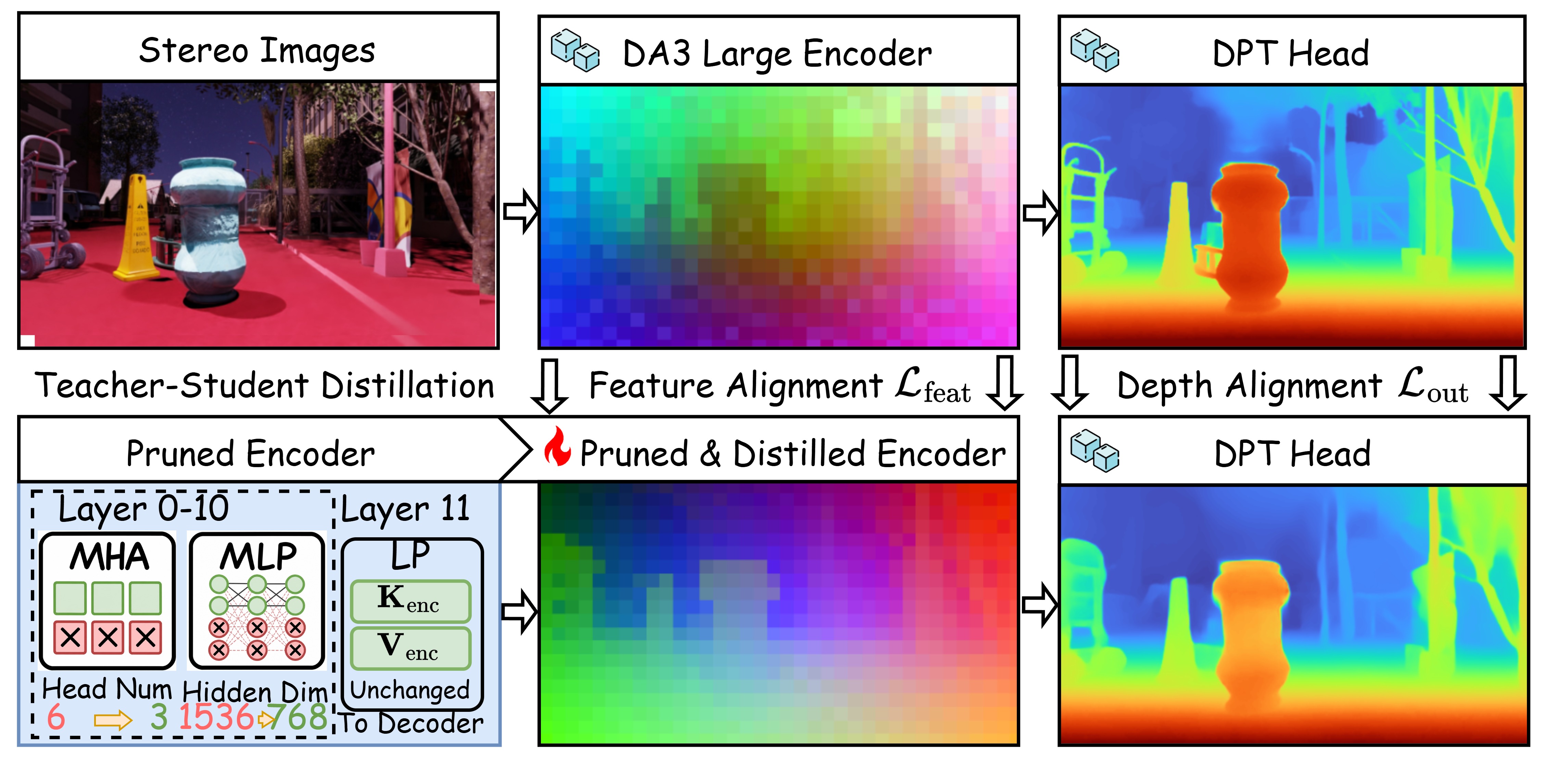}
    \vspace{-1em}
    \caption{Pruning and teacher-student distillation of the Scene Representor.}
    \label{Fig: Distill}
    \vspace{-1em}
\end{figure}
To preserve the alternating within-view and cross-view attention structure of the Scene Representor, we retain all $12$ Transformer layers and the original embedding dimension $D=384$, while reducing the internal widths of layers $0$–$10$. The linear projections in the final Transformer layer (indexed $11$) remain unchanged to preserve the dimensional compatibility of the key–value representations with the Trajectory Decoder.
Specifically, in Transformer layers $0$–$10$, both the number of attention heads $h$ and the MLP hidden dimension $D_m$ are halved, as illustrated in Fig.~\ref{Fig: Distill}, resulting in an approximately $50\%$ reduction in the dominant computational cost of each layer.
This follows from the per-layer computational complexity of $\mathcal{O}\left(\rho D h d_h + \rho^2 h d_h + \rho D D_m\right)$, where $\rho$ denotes the number of tokens and $d_h=64$ is the per-head dimension, consistent with the DA3-Small encoder.

Since this pruning strategy only reduces the width of each layer while preserving the overall architecture, the pruned encoder is a sub-network of the Scene Representor.
It can therefore be directly initialized with the pretrained weights of the retained attention heads and MLP channels of DA3-Small, thereby inheriting a reasonable geometry-aware perception capability at initialization, as shown in Fig.~\ref{Fig: Distill}.
Furthermore, we reduce the base input resolution from $504$, as adopted in the pretrained DA3 with varying aspect ratios, to $336$ and employ knowledge distillation to mitigate the resulting degradation in geometric perception performance.
\begin{figure}
\centering	
\subfigure[Token-wise features from the pruned and distilled Scene Representor for the simulation (Fig.~\ref{Fig: NetworkArchitecture}{\color{blue}(a)}) and real-world (Fig.~\ref{Fig: NetworkArchitecture}{\color{blue}(b)}) images, with cosine similarity maps between their corresponding scene tokens.]{
    \includegraphics[width=0.95\linewidth]{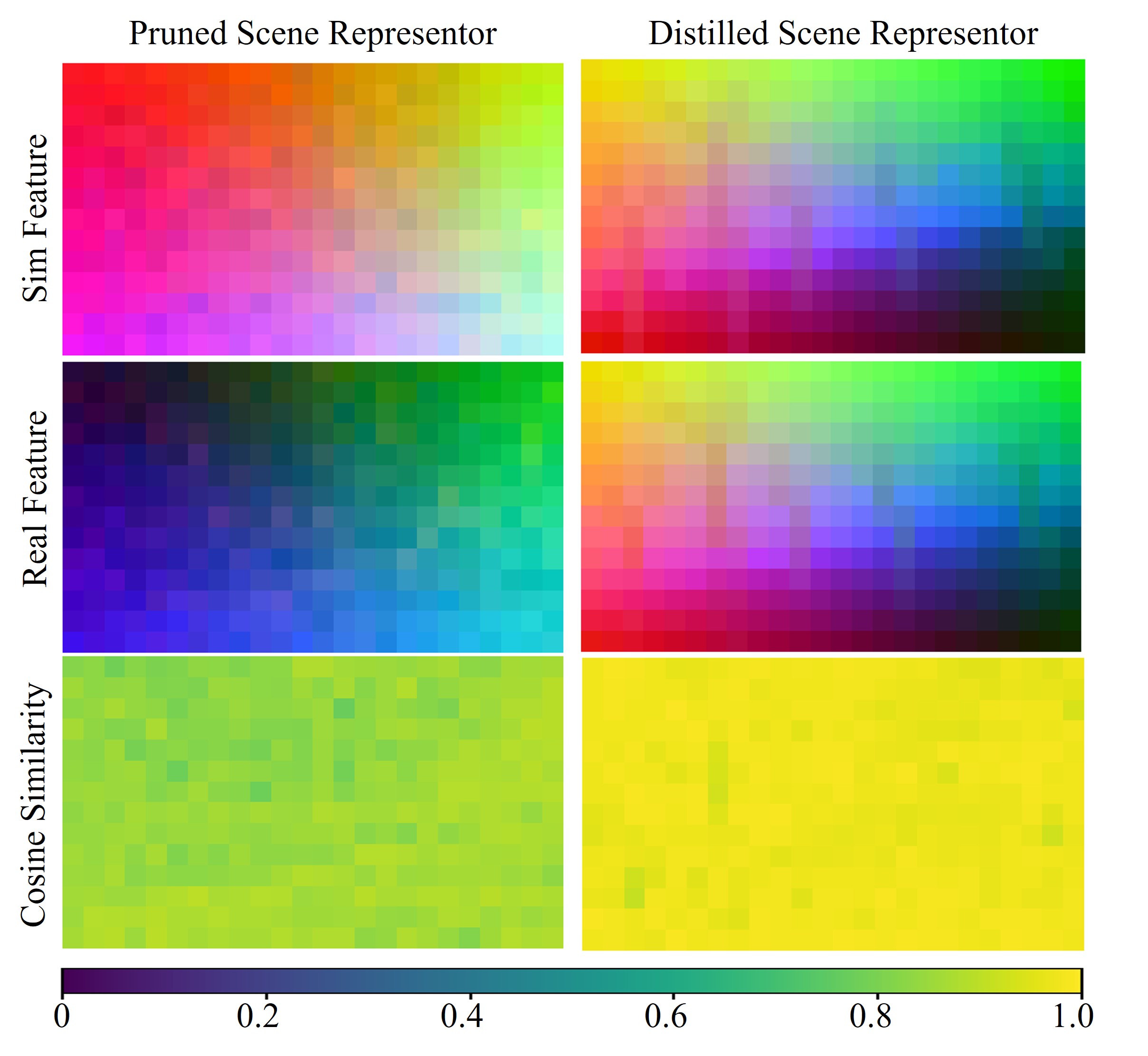}
    \label{Fig: Sim2RealFeatMapDistill}
}
\subfigure[Token-wise cosine similarity distribution]{
    \includegraphics[width=\linewidth]{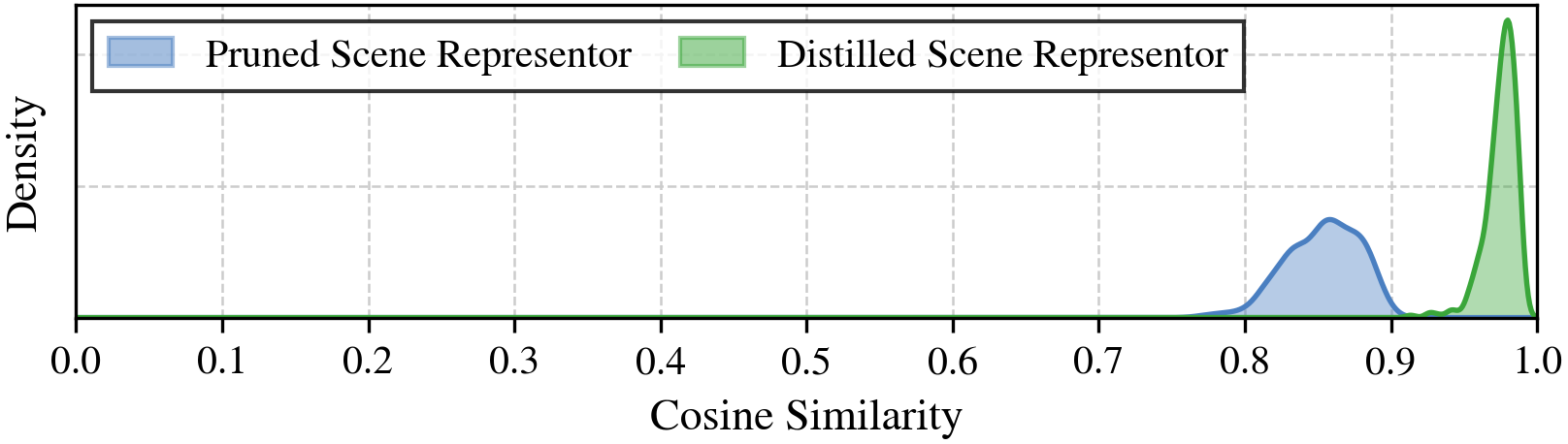}
    \label{Fig: Sim2RealSimilarityDisDistill}
}
\caption{Sim-to-real transfer of the distilled Scene Representor.}
\label{Fig: Sim2RealDistill}
\vspace{-1em}
\end{figure}
\begin{figure}[!t]
    \centering
    \subfigure[Forest]{\includegraphics[width=0.49\linewidth]{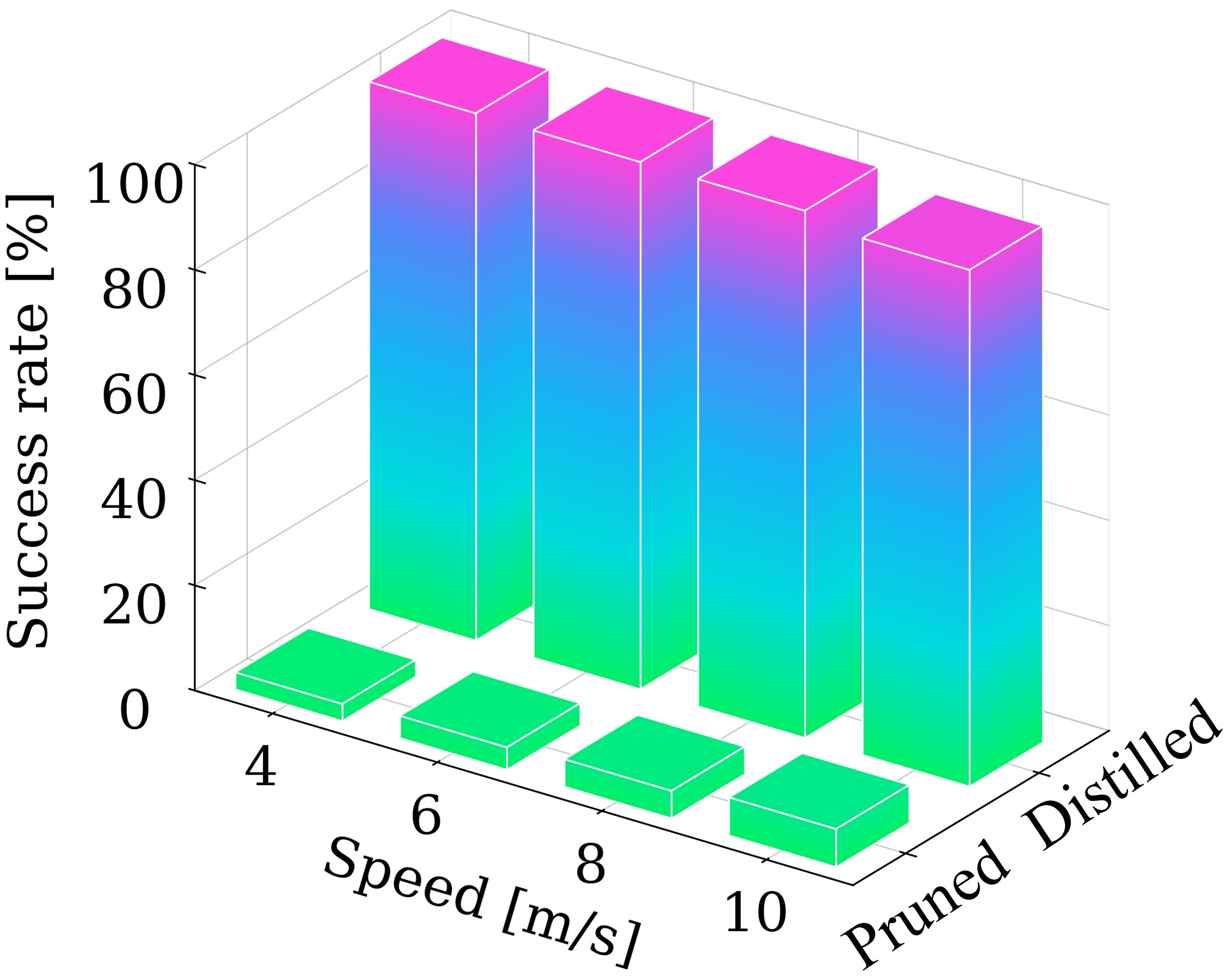}}\hfill
    \subfigure[Industrial]{\includegraphics[width=0.49\linewidth]{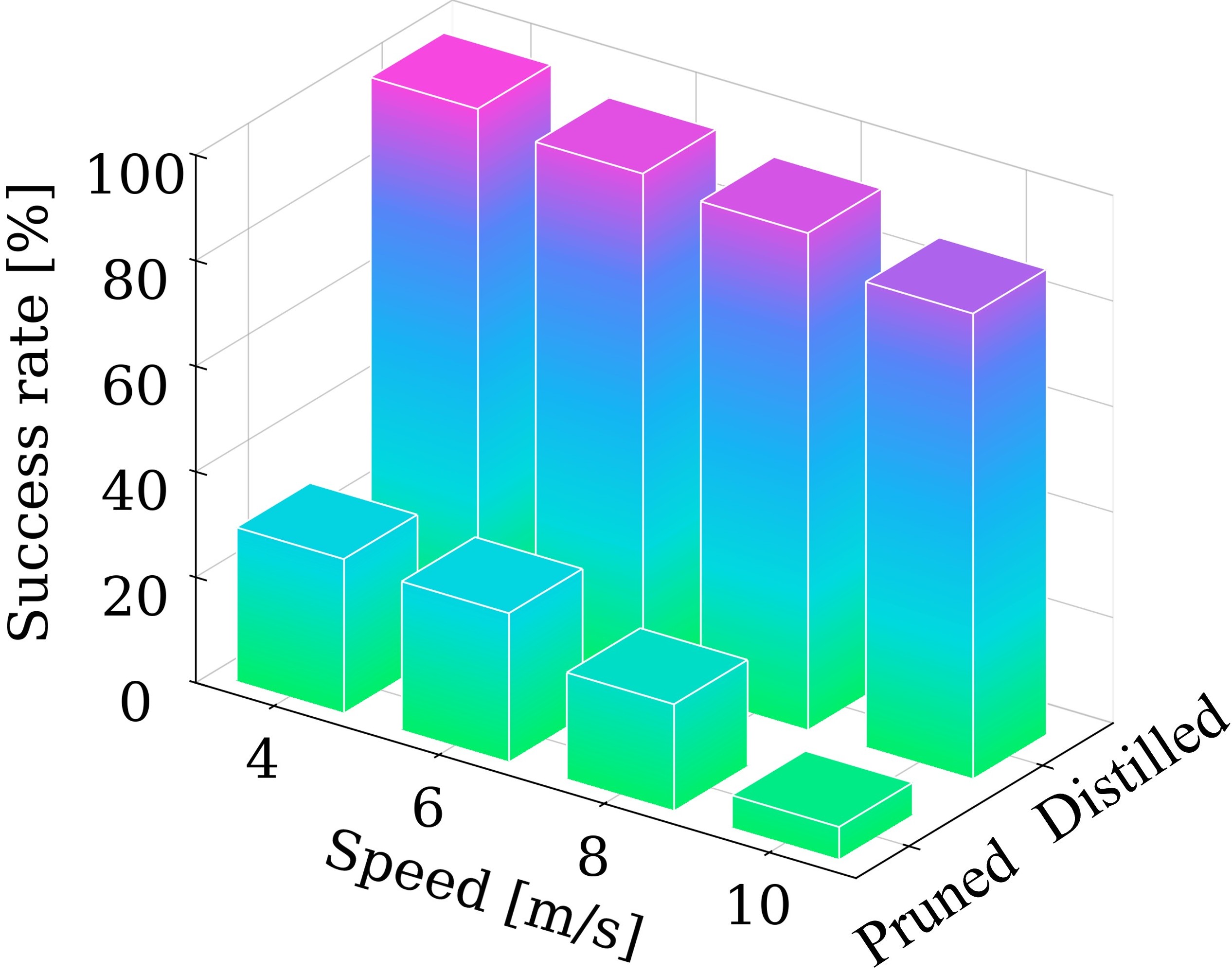}}
    \caption{Success rates of Stereo-Planner with the pruned and the pruned-and-distilled Scene Representors in the forest and industrial environments.}
    \label{Fig: DistillSuccess}
    \vspace{-1em}
\end{figure}
\begin{figure*}[!t]
    \centering
    \includegraphics[width=\linewidth]{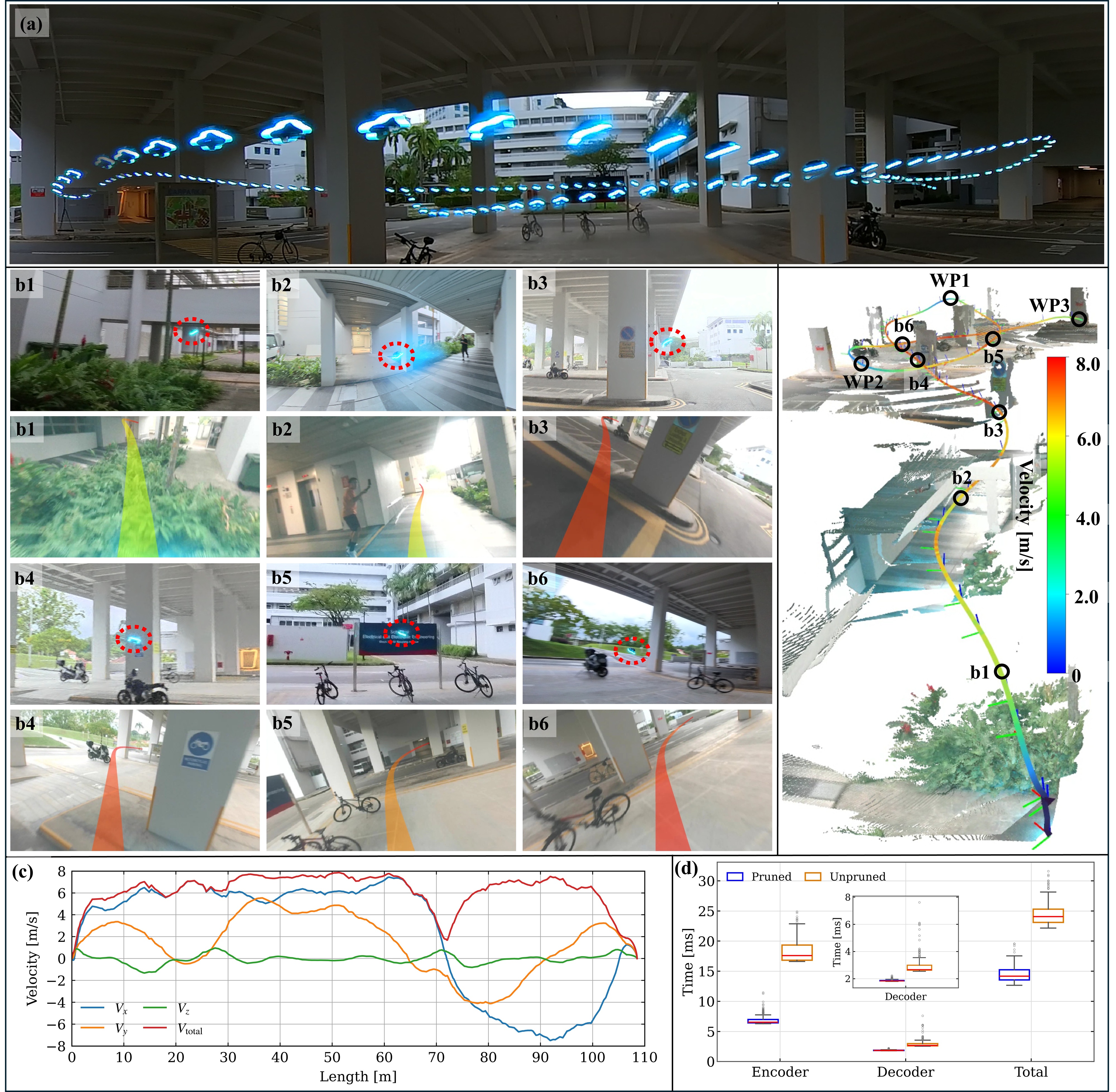}
    \caption{Autonomous flight in a vehicle loading bay using Stereo-Planner with the distilled Scene Representor. (a) Overview of the executed flight trajectory. (b) Executed flight trajectory color-coded by flight speed, passing through waypoints WP1–WP3. (c) Speed profiles along each axis and the total speed. (d) Onboard processing time of the Scene Representor (Encoder), the Trajectory Decoder (Decoder), and the overall vision-to-action pipeline (Total) on the NVIDIA Orin NX with the distilled (Pruned) and original (Unpruned) Scene Representors.}
    \label{Fig: DistillFlight}
\end{figure*}
\subsection{Distillation}
To recover the geometric perception capability lost through pruning, we adopt knowledge distillation, in which DA3-Large serves as the teacher network to supervise the training of the pruned encoder (student network).
Specifically, the student is supervised by a combination of an output distillation loss $\mathcal{L}_{\rm{out}}$ and an intermediate feature distillation loss $\mathcal{L}_{\rm{feat}}$:
\begin{equation}
\mathcal{L} = \mathcal{L}_{\rm{out}} + \lambda_{\rm{feat}}\mathcal{L}_{\rm{feat}}
\end{equation}
where $\lambda_{\rm{feat}}$ is the balancing coefficient, empirically set to $0.5$.

The output distillation loss $\mathcal{L}_{\rm{out}}$ supervises the depth maps predicted by the student.
To remove the scale ambiguity, the depth maps predicted by the student and the teacher are normalized by the per-sample median $Z_m$ of the teacher depth, i.e., $\hat Z_s = Z_s/Z_m$ and $\hat Z_t = Z_t/Z_m$.
\begin{equation}
\mathcal{L}_{\rm{out}} = \frac{1}{|\Omega|}\sum_{p\in\Omega} w_p \left| \hat Z_s(p) - \hat Z_t(p) \right|,\quad \Omega = \left\{ p \,\middle|\, \hat Z_t(p) \le 20 \right\}
\end{equation}
where $p$ indexes the pixels, and the distant or sky pixels with normalized depth larger than $20$ are excluded.
The weight $w_p$ is the depth confidence predicted by the teacher, normalized to have a mean of $1$ and clipped at an upper bound of $10$.

The intermediate feature distillation loss transfers the geometry-aware representations of the teacher to the student encoder.
For a set $\mathbb{P}$ of teacher-student layer pairs, the student patch tokens after the final LayerNorm are mapped to the teacher dimension through a learnable asymmetric projection ${\bf{W}}_k \in \mathbb{R}^{384\times1024}$ without bias.
\begin{equation}
\begin{aligned}
&\mathcal{L}_{\rm{feat}} = \frac{1}{|\mathbb{P}|}\sum_{(k,l)\in\mathbb{P}} \frac{1}{\rho}\sum_{j=1}^{\rho} \mathbf{1}\left[ c_j^{k,l} < \alpha \right]\left( 1 - c_j^{k,l} \right)\\
&c_j^{k,l} = \frac{{\bf{z}}_{s,j}^{k}{\bf{W}}_k {\bf{z}}_{t,j}^{l,\top}}{\left\| {\bf{z}}_{s,j}^{k}{\bf{W}}_k \right\|_2 \left\| {\bf{z}}_{t,j}^{l} \right\|_2}
\end{aligned}
\end{equation}
where ${\bf{z}}_{s,j}^{k}$ and ${\bf{z}}_{t,j}^{l}$ denote the $j$-th patch token from the $k$-th student layer and the $l$-th teacher layer, respectively; $c_j^{k,l}$ denotes the cosine similarity between the projected student token and the corresponding teacher token; $\mathbf{1}[\cdot]$ is the indicator function; and $\alpha=0.9$ is the margin, such that only tokens whose cosine similarity falls below $\alpha$ contribute to the loss.

\subsection{Training Dataset}
Since DA3\cite{depthanything3} is a strong vision grounding foundation model with powerful geometric perception capability acquired from large-scale and diverse pretraining data, the purpose of distillation is to preserve this general-purpose capability in the pruned encoder rather than to specialize it to the planning task.
To this end, the distillation is performed on the large-scale synthetic stereo dataset released with FoundationStereo\cite{FoundationStereo}, which provides a million stereo image pairs with substantial scale and diversity in scene layouts, object categories, materials, and lighting conditions.
Notably, the student network is not distilled on the Random Obstacles scenes described in Section~\ref{Sec: Training Data Collection}, to avoid overfitting to the specific obstacle configurations used for evaluating Stereo-Planner.

\subsection{Benchmark}
To verify whether distillation can restore the geometric perception capability of the pruned student network, we evaluate the scene representations extracted by the pruned and distilled Scene Representor on the simulated and real-world images shown in Fig.~\ref{Fig: NetworkArchitecture}{\color{blue}(a)} and Fig.~\ref{Fig: NetworkArchitecture}{\color{blue}(b)}.
As shown in Fig.~\ref{Fig: Sim2RealFeatMapDistill}, the distilled encoder clearly perceives the spatial geometric structure of the scene, and extracts similar geometric representations from the simulated and real-world images that differ in visual appearance but share the same geometric structure, indicating a substantial improvement in the geometric representation capability of the pruned encoder.
Furthermore, as shown in Fig.~\ref{Fig: Sim2RealSimilarityDisDistill}, the token-wise cosine similarities between the simulated and real-world scene tokens of the distilled Scene Representor are predominantly distributed above $0.9$, which is comparable to those of the original DA3-Small encoder in Fig.~\ref{Fig: Sim2RealSimilarityDis}.
To further verify whether the distilled Scene Representor provides Stereo-Planner with geometric perception, we evaluate Stereo-Planner equipped with the distilled Scene Representor in the forest and industrial simulation environments shown in Fig.~\ref{Fig: SimScene}{\color{blue}(b)} and Fig.~\ref{Fig: SimScene}{\color{blue}(c)}, following the protocol in Section~\ref{Sec: Experimental Evaluations and Validations} at commanded speeds of $4$, $6$, $8$, and $10\,\mathrm{m/s}$.
As shown in Fig.~\ref{Fig: DistillSuccess}, Stereo-Planner with the distilled Scene Representor achieves success rates comparable to those obtained with the original Scene Representor, and substantially outperforms its counterpart equipped with the pruned but undistilled Scene Representor.
These results demonstrate that the distilled Scene Representor preserves the cross-domain consistency of the geometric representations learned by the original encoder.

\subsection{Flight with Distilled Scene Representor}
To demonstrate the practical deployability of the distilled Scene Representor, we replace the original Scene Representor in Stereo-Planner with its distilled counterpart and deploy the resulting framework on the Stereo-MAV platform shown in Fig.~\ref{Fig: Platform}.
In this experiment, Stereo-MAV performs autonomous flight in a previously unseen vehicle loading bay, as shown in Fig.~\ref{Fig: DistillFlight}{\color{blue}(a)}.
The quadrotor successfully navigates through the waypoints (WP1-WP3 in Fig.~\ref{Fig: DistillFlight}{\color{blue}(b)}), executing agile and aggressive maneuvers while avoiding obstacles, and reaches a maximum speed of $7.9\,\mathrm{m/s}$, as shown in Fig.~\ref{Fig: DistillFlight}{\color{blue}(c)}.
Fig.~\ref{Fig: DistillFlight}{\color{blue}(d)} reports the onboard processing time of Stereo-Planner with the distilled Scene Representor.
With TensorRT acceleration, the average inference time of the distilled Scene Representor is $6.85\,\mathrm{ms}$, a $62.3\%$ reduction compared with $18.18\,\mathrm{ms}$ for the original Scene Representor.
Moreover, since the distilled Scene Representor adopts a base resolution of $336$ instead of the base resolution of $504$ used in DA3 pretraining, fewer scene tokens are passed to the Trajectory Decoder, reducing the trajectory decoding time from $2.87\,\mathrm{ms}$ to $1.87\,\mathrm{ms}$.
Consequently, the average vision-to-action processing time of Stereo-Planner is reduced from $24.27\,\mathrm{ms}$ to $14.44\,\mathrm{ms}$, a $40.6\%$ reduction, which further improves the real-time performance of quadrotor onboard navigation.
\bibliographystyle{IEEEtran}
\bibliography{IEEEabrv,main}
\end{document}